\documentclass{article}
\usepackage{algorithm}
\usepackage{algpseudocode}
\usepackage{booktabs}
\usepackage{graphicx}
\usepackage{multirow}
\usepackage{caption}
\usepackage{amsmath}
\usepackage{amssymb}

\algrenewcommand\algorithmicrequire{\textbf{Input:}}
\algrenewcommand\algorithmicensure{\textbf{Output:}}
\usepackage[preprint]{corl_2026} 

\title{Goal Sets, Not Goal States: Queryable Robot Goals through Goal-Set Hindsight Relabeling}

\author{
  Carlos Vélez García\\
  INESCOP\\
  Elda, Alicante, Spain\\
  \texttt{cvelez@inescop.es} \\
  \And
  Miguel Cazorla \\
  University of Alicante \\
  Alicante, Spain \\
  \And
  Jorge Pomares \\
  University of Alicante \\
  Alicante, Spain \\
}

\begin{document}
\maketitle


\begin{abstract}
Hindsight relabeling usually turns achieved future states into exact goals, which can overconstrain offline robot learning when task success depends only on a subset of the state. We propose Goal-Set Hindsight Relabeling (GS-HER), a predicate-level generalization of HER in which achieved states certify query-defined goal sets rather than singleton goal states. A binary query specifies which variables define success, making the goal predicate an inference-time input while leaving the underlying offline GCRL algorithm unchanged. Across OGBench tasks and five offline goal-conditioned learners, GS-HER improves performance when full-state goals are bottlenecked by nuisance dimensions and turns hindsight relabeling into a reusable goal interface: one checkpoint can answer multiple robot goal predicates without retraining.\footnote{Code available at \url{https://github.com/cvg25/gs_her}.}
\end{abstract}

\section{Introduction}
\label{sec:introduction}

Offline robot learning aims to turn previously collected experience---demonstrations, play data, teleoperation logs, or exploratory trajectories---into useful control policies without costly online trial and error. A particularly attractive formulation is offline goal-conditioned reinforcement learning (GCRL), where agents learn from reward-free trajectories to reach user-specified goals. Hindsight Experience Replay (HER) provides the central mechanism: states achieved later in a trajectory are replayed as goals for earlier states~\citep{andrychowicz2017hindsight}. Together with universal value function approximators~\citep{schaul2015universal}, this enables a single goal-conditioned model to represent many goal-reaching problems.

Recent offline GCRL benchmarks such as OGBench often instantiate this idea with the full state space as the goal space, so every future state can become a relabeled goal~\citep{park2025ogbench}. This makes relabeling unsupervised and domain-agnostic, but it also imposes a strong semantic assumption: success means reproducing the entire future state. In manipulation, this can be overly restrictive. A cube-placement task may require matching the cube position while ignoring robot configuration, gripper state, velocities, and other nuisance variables. Full-state HER therefore turns task-irrelevant coordinates into artificial constraints.

We view this as an objective-specification problem. Standard HER treats an achieved future state \(s_{t+k}\) as a singleton goal \(g=s_{t+k}\). Our key observation is:
\begin{quote}
\emph{An achieved state certifies reachability not only of itself, but of every goal set that contains it.}
\end{quote}
Thus, if a trajectory reaches \(s_{t+k}\), it provides supervision for any goal set \(\mathcal{G}\) such that \(s_{t+k}\in\mathcal{G}\). Standard HER is the special case \(\mathcal{G}=\{s_{t+k}\}\). Many robot tasks instead correspond to larger equivalence classes: all states with the same object position, drawer state, button configuration, end-effector pose, or other task-relevant variables.

We propose Goal-Set Hindsight Relabeling (GS-HER), a query-conditioned, predicate-level generalization of HER that turns achieved states into certificates for queryable goal sets. In vector-state environments, a binary query \(q\in\{0,1\}^D\) specifies which state variables define success. Active coordinates define the goal predicate; inactive coordinates are nuisance variables for that query. GS-HER therefore replaces exact-state success with a query-conditioned goal-set predicate while leaving the underlying offline GCRL algorithm unchanged.

\begin{figure*}[t]
    \centering
    \includegraphics[width=\textwidth]{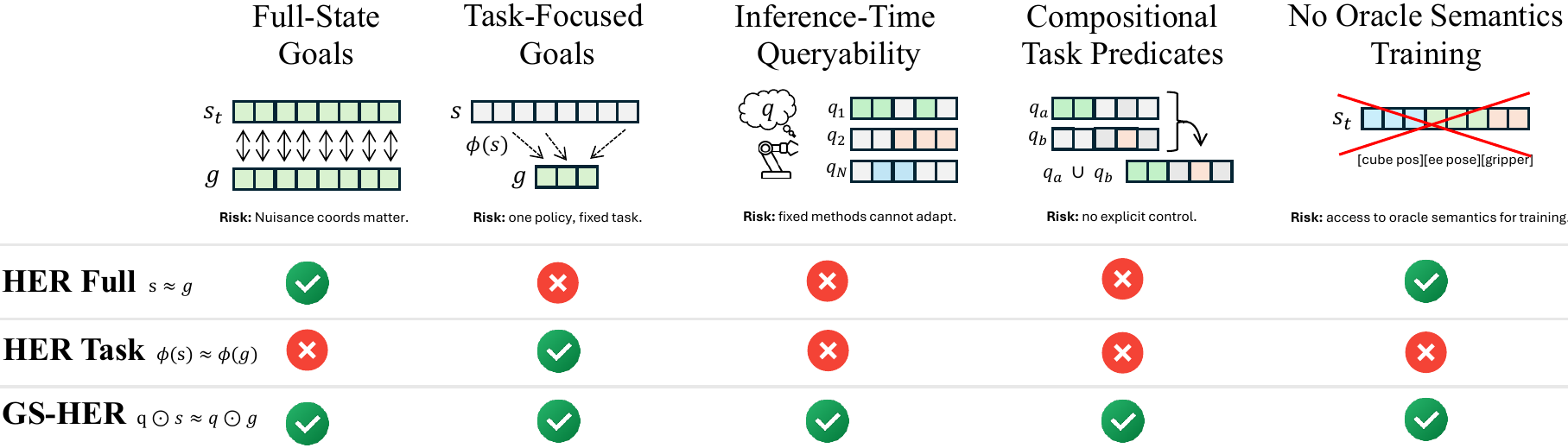}
    \caption{
    \textbf{From many goal states to many goal predicates.}
    HER-Full supports annotation-free relabeling but fixes success to full-state matching. HER-Task focuses learning through a fixed oracle projection $\phi$, but fixes the task semantics before training. GS-HER avoids oracle task projections while conditioning on a query $q$, allowing the same model to recover full-state goals, task-focused goals, and compositional predicates at inference time.
    }
    \label{fig:methods_comparison}
\end{figure*}

This turns hindsight relabeling into a reusable goal interface. Instead of training one policy for full-state matching or one policy for a fixed task projection, GS-HER trains a single checkpoint that can be asked different questions at inference time: match the cube position, match the end-effector pose, satisfy both object position and orientation, or recover full-state reaching by activating all coordinates. In this view, the reference state specifies the desired values, while the query specifies what counts as success. Figure~\ref{fig:methods_comparison} summarizes this distinction. A downstream user, planner, or operator can therefore change the task semantics without recollecting data, redefining rewards, or retraining a separate model.

Our contributions are:
\begin{enumerate}
    \item We identify an objective mismatch in full-state offline GCRL: exact-state hindsight relabeling can overconstrain robotic tasks whose success criteria depend only on task-relevant projections of the state.
    \item We formalize Goal-Set Hindsight Relabeling as a predicate-level generalization of HER in which achieved states certify reachability of goal sets rather than only singleton goal states.
    \item We instantiate GS-HER as a query-conditioned relabeling wrapper for vector-state environments, enabling existing offline GCRL algorithms to learn queryable goal semantics without changing their learning objectives.
    \item We show that GS-HER improves OGBench performance when full-state goals are bottlenecked by nuisance dimensions, and enables one trained model to answer multiple goal predicates at inference time without retraining.
\end{enumerate}

\section{Related Work}
\label{sec:related_work}

Goal-conditioned reinforcement learning learns policies and value functions conditioned on desired outcomes, commonly using hindsight relabeling to treat achieved future states as goals~\citep{kaelbling1993learning,schaul2015universal,andrychowicz2017hindsight}. Although HER is general enough to support projected or predicate-defined goals, the goal semantics are typically fixed before training. A policy trained for object-position goals, for example, does not expose an interface for switching at test time to object orientation, end-effector pose, gripper state, or their compositions without redefining the goal space.

Recent offline GCRL methods learn from fixed, reward-free datasets using behavioral cloning, value learning, contrastive objectives, or hierarchical structure~\citep{lynch2020learning,ghosh2020learningreachgoalsiterated,ding2019goal,kostrikov2021offline,eysenbach2022contrastive,park2023hiql,park2025ogbench}. Benchmarks such as OGBench often instantiate this setting with full-state goals, making hindsight relabeling domain-agnostic but tying success to exact-state reachability. In domains, such as robotic manipulation, this can overconstrain learning because task success may depend on object state while robot configuration, velocities, and other coordinates are nuisance variables.

Prior work has also explored alternative goal interfaces, including object-coordinate goals, goal images, learned latent goals, language-conditioned policies, and hierarchical skill spaces~\citep{nair2018visual,pong2018temporal,florensa2018automatic,pong2019skew,gehring2021hierarchical,hafner2022director}. These methods improve flexibility, but they either commit to a fixed success semantics during training, require external task supervision, or introduce additional hierarchy or world-model machinery. GS-HER is complementary: it changes the semantic object produced by hindsight relabeling. An achieved state supervises not only a singleton goal state, but any query-defined goal set that contains it, enabling existing offline GCRL learners to expose goal semantics as an inference-time query.  We provide an extended discussion of related work in Appendix~\ref{app:extended_related_work}.

\section{Goal-Set Hindsight Relabeling}
\label{sec:method}

We consider an offline dataset of reward-free trajectories
\[
\mathcal{D}
=
\{\tau^{(n)}=(s^{(n)}_0,a^{(n)}_0,s^{(n)}_1,\ldots,s^{(n)}_{T_n})\}_{n=1}^N,
\]
with states \(s\in\mathcal{S}\subseteq\mathbb{R}^D\) and actions
\(a\in\mathcal{A}\). Standard hindsight relabeling turns states observed
in the dataset into goals for earlier transitions. In most offline
goal-conditioned pipelines, this is instantiated as an exact goal-state
objective: a relabeled transition is successful only if the agent reaches
the complete reference state \(g\). GS-HER generalizes this operation by
treating a relabeled state as evidence for a family of goal sets rather
than only for a singleton goal state.

\subsection{Goal sets as query-defined equivalence classes}

The central object in GS-HER is a goal query. A query specifies which
state variables define success:
\[
q\in\{0,1\}^D,
\qquad
\|q\|_0\geq 1,
\]
where \(q_i=1\) marks coordinate \(i\) as active in the goal predicate and
\(q_i=0\) marks it as a nuisance variable for the current task. Given a
reference state \(g\), the query induces an equivalence relation
\[
s \equiv_q g
\quad\Longleftrightarrow\quad
q\odot s = q\odot g,
\]
and therefore the goal set
\[
\mathcal{G}(g,q)
=
\{s'\in\mathcal{S}: q\odot s' = q\odot g\}.
\]
Thus, reaching \(g\) under query \(q\) does not mean reproducing the
entire reference state; it means matching only the coordinates declared
active by the query.

For continuous control, exact equality is replaced by a
query-conditioned discrepancy. We use the normalized squared distance
\[
d_q(s',g)
=
\frac{1}{\|q\|_0}
\sum_{i=1}^D q_i (s'_i-g_i)^2,
\]
with the associated tolerance-relaxed goal set
\[
\mathcal{G}_{\epsilon}(g,q)
=
\{s'\in\mathcal{S}: d_q(s',g)\leq \epsilon_q\}.
\]
The success predicate becomes
\[
c_{g,q}(s')
=
\mathbb{I}\!\left[d_q(s',g)\leq \epsilon_q\right].
\]
Other coordinate-wise metrics can be substituted when the state contains
structured variables such as angles or rotations; the only requirement is
that \(c_{g,q}\) defines membership in the query-specified goal set.

This formulation recovers standard hindsight relabeling as the special
case \(q=\mathbf{1}\). When all coordinates are active,
\(\mathcal{G}(g,\mathbf{1})=\{g\}\) up to the tolerance, and GS-HER
reduces to full-state HER at the predicate level. When only a subset of
coordinates is active, the same relabeled reference state supervises a
projected goal-reaching problem over the same offline data.

\subsection{Query-conditioned goal interface}

GS-HER is implementation-agnostic with respect to the underlying offline
goal-conditioned learner. The base algorithm may learn a policy, value
function, critic, contrastive score, distance estimate, or behavioral
cloning objective. GS-HER only changes the goal-conditioning interface
and the success predicate used for relabeling.

To expose the query to the learner, we define a query-conditioned
conditioning input
\[
\psi_{\mathrm{GS}}(s,g,q)
=
[\bar{s}_q,\bar{g}_q,q],
\]
where
\[
\bar{s}_q
=
q\odot s + (1-q)\odot e_s,
\qquad
\bar{g}_q
=
q\odot g + (1-q)\odot e_g.
\]
Here \(e_s,e_g\in\mathbb{R}^D\) are nuisance-coordinate embeddings used
only in inactive coordinates. They may be learned parameters, fixed
constants, or samples from a state distribution. Their role is not to
corrupt the input, but to remove task-irrelevant values from the
goal-conditioning channel while the query \(q\) explicitly states which
coordinates define success.

The raw state remains available to the base learner in the usual way.
For example, GS-HER-conditioned objects can be written as
\[
\pi(a\mid s,\psi_{\mathrm{GS}}(s,g,q)),
\qquad
Q(s,a,\psi_{\mathrm{GS}}(s,g,q)),
\qquad
V(s,\psi_{\mathrm{GS}}(s,g,q)).
\]
The additional \(\bar{s}_q\) term in \(\psi_{\mathrm{GS}}\) should be
understood as part of the goal-comparison interface: it presents the
current and reference states under the same active coordinates, while
the unmasked state \(s\) remains the control input. This makes GS-HER
directly comparable to common implementations of HER-Full and HER-Task,
which condition on state--goal or projected state--projected goal pairs.

\subsection{GS-HER as a relabeling wrapper}

GS-HER wraps an existing goal-conditioned learner. Given a transition
\((s_t,a_t,s_{t+1})\), the base relabeling sampler first chooses a
reference state \(g\). This sampler is not prescribed by GS-HER: it may
select the current state, a future state from the same trajectory, or a
random state from the dataset, depending on the requirements of the
underlying algorithm. GS-HER then samples a query
\[
q\sim p_Q(q),
\qquad
q\in\{0,1\}^D,\quad \|q\|_0\geq 1,
\]
constructs \(\psi_{\mathrm{GS}}(s_t,g,q)\), and evaluates success using
\(c_{g,q}\) instead of exact full-state matching.

The learning objective of the base algorithm is otherwise unchanged. If
the base method uses Bellman or expectile updates, GS-HER changes only
the reward or success labels through \(c_{g,q}\). If the base method is
behavioral cloning, contrastive learning, or distance learning, GS-HER
provides the same relabeled reference through the query-conditioned
interface \(\psi_{\mathrm{GS}}\). Thus, GS-HER changes what each relabeled
transition means, not how the base learner optimizes its objective. Algorithm~\ref{alg:gs-her} summarizes GS-HER as a wrapper around a base goal-conditioned learner: the goal sampler and learning objective are inherited from the base algorithm, while GS-HER replaces the relabeled goal semantics with a query-conditioned goal set.

\begin{algorithm}[t]
\caption{Goal-Set Hindsight Relabeling (GS-HER)}
\label{alg:gs-her}
\begin{algorithmic}[1]
\Require Dataset \(\mathcal{D}\), base goal-conditioned learner \(\mathrm{Alg}\), goal sampler \(p_G\), query distribution \(p_Q\), nuisance embeddings \(e_s,e_g\)
\Ensure Trained goal-conditioned model
\State Initialize \(\mathrm{Alg}\)
\Repeat
    \State Sample trajectory \(\tau=(s_0,a_0,\ldots,s_T)\sim\mathcal{D}\)
    \State Sample transition \((s_t,a_t,s_{t+1})\) from \(\tau\)
    \State Sample reference state \(g\sim p_G(\cdot\mid \tau,t,\mathcal{D})\)
    \State Sample query \(q\sim p_Q(q)\), with \(\|q\|_0\geq 1\)
    \State Construct query-conditioned input:
    \[
    \psi_{\mathrm{GS}}(s_t,g,q)
    =
    [q\odot s_t+(1-q)\odot e_s,\;
      q\odot g+(1-q)\odot e_g,\;
      q]
    \]
    \State Define goal-set success:
    \[
    c_{g,q}(s')
    =
    \mathbb{I}
    \left[
    \frac{1}{\|q\|_0}
    \sum_i q_i(s'_i-g_i)^2
    \leq \epsilon_q
    \right]
    \]
    \State Relabel the transition with \((\psi_{\mathrm{GS}}(s_t,g,q),c_{g,q})\)
    \State Update \(\mathrm{Alg}\) using its original objective
\Until{convergence}
\end{algorithmic}
\end{algorithm}

The query distribution \(p_Q\) determines the family of goal sets that
the trained model can answer. Full-state queries recover standard
full-state HER. A fixed task query recovers conventional projected-goal
HER at the predicate level. Random, blockwise, or semantic queries train
a reusable interface over many possible goal predicates. In our
experiments, the main domain-agnostic variant uses blockwise queries,
while semantic queries are used as an oracle-query variant when a
state-factorization is available. Query generation details are given in Appendix~\ref{app:data_sampling}.

\subsection{Inference with goal queries}

At inference time, the query becomes the task interface. Given a desired
reference state \(g^\star\) and a task query \(q^\star\), we construct
\[
\psi_{\mathrm{GS}}(s,g^\star,q^\star)
=
[
q^\star\odot s+(1-q^\star)\odot e_s,\;
q^\star\odot g^\star+(1-q^\star)\odot e_g,\;
q^\star
],
\]
and evaluate the trained policy, value function, critic, or score using
this conditioning input.

Changing \(q^\star\) changes the definition of success without changing
the model weights. A cube-position task activates only cube-position
coordinates; a full-state task uses \(q^\star=\mathbf{1}\); and a
stricter object-configuration task may activate both position and
orientation. The same checkpoint can therefore be queried with different
goal predicates at test time, turning hindsight relabeling from a
single-predicate training device into a reusable goal interface.



\section{Experiments}
\label{sec:experiments}

We evaluate GS-HER along four axes: whether query-conditioned relabeling improves full-state task performance, whether a single trained model can answer multiple goal predicates at inference time, and whether the gains arise when full-state goals are overconstrained by nuisance dimensions. We also test whether these effects are orthogonal to the underlying learner.

\subsection{Experimental Setup}

\paragraph{Environments.}
We evaluate on state-based OGBench tasks~\citep{park2025ogbench}. Our manipulation suite includes \texttt{cube-single-play-v0}, \texttt{cube-single-noisy-v0}, \texttt{cube-double-play-v0}, \texttt{cube-double-noisy-v0}, and \texttt{scene-play-v0}. These domains expose the failure mode studied in this paper: official success depends on object configurations, while the full state also contains robot configuration, gripper state, velocities, and other nuisance variables. We also evaluate \texttt{pointmaze-medium-navigate-v0}, \texttt{pointmaze-large-navigate-v0}, and \texttt{antmaze-medium-navigate-v0}. PointMaze is a low-nuisance control where the state already matches the goal predicate; AntMaze tests whether GS-HER can ignore high-dimensional locomotion variables when the task is navigation. Task visualizations are shown in Appendix \ref{app:task_visualizations}

\paragraph{Relabeling schemes.}
We compare four relabeling strategies. \textbf{HER-Full} is the standard domain-agnostic protocol: achieved future states are relabeled as exact full-state goals, so success requires matching every coordinate. \textbf{HER-Task} is an oracle fixed-projection baseline: goals contain only the coordinates used by the official task predicate, but the projection is chosen before training and therefore defines a single task semantics. \textbf{GS-HER Blockwise} is our default domain-agnostic variant: each relabeled goal is paired with a query sampled from full-state, random-coordinate, and contiguous coordinate-block queries, requiring no semantic state annotations. \textbf{GS-HER Semantic} is an oracle-query variant in which queries activate predefined physical factors, such as object position, object orientation, robot pose, or gripper state. Comparing Blockwise and Semantic separates query-conditioned relabeling from the additional benefit of privileged state factorization.

\paragraph{Base learners.}
GS-HER is a relabeling operator, so we evaluate it as a wrapper around five offline GCRL algorithms: goal-conditioned behavioral cloning (GCBC)~\citep{lynch2020learning,ghosh2020learningreachgoalsiterated}, contrastive RL (CRL)~\citep{eysenbach2022contrastive}, goal-conditioned implicit \(\{V,Q\}\)-learning (GCIVL and GCIQL)~\citep{kostrikov2021offline,park2023hiql}, and Goal-Conditioned Distance Learning (GCDL), a non-discounted diagnostic distance-learning variant. These cover imitation-based, contrastive, value-based, and distance-based goal-conditioned learning. In all cases, GS-HER changes only the relabeled goal representation and success predicate; the base objective, architecture class, and optimization protocol are otherwise unchanged. Thus, differences between HER-Full, HER-Task, and GS-HER isolate the effect of goal semantics rather than changes to the underlying learner. Additional details are provided in Appendix~\ref{app:base_algorithms}.

\subsection{Full-State Hindsight Relabeling Bottlenecks Official Task Performance}
\label{sec:exp_official_task_performance}

\begin{figure*}[b]
    \centering
    \includegraphics[width=\textwidth]{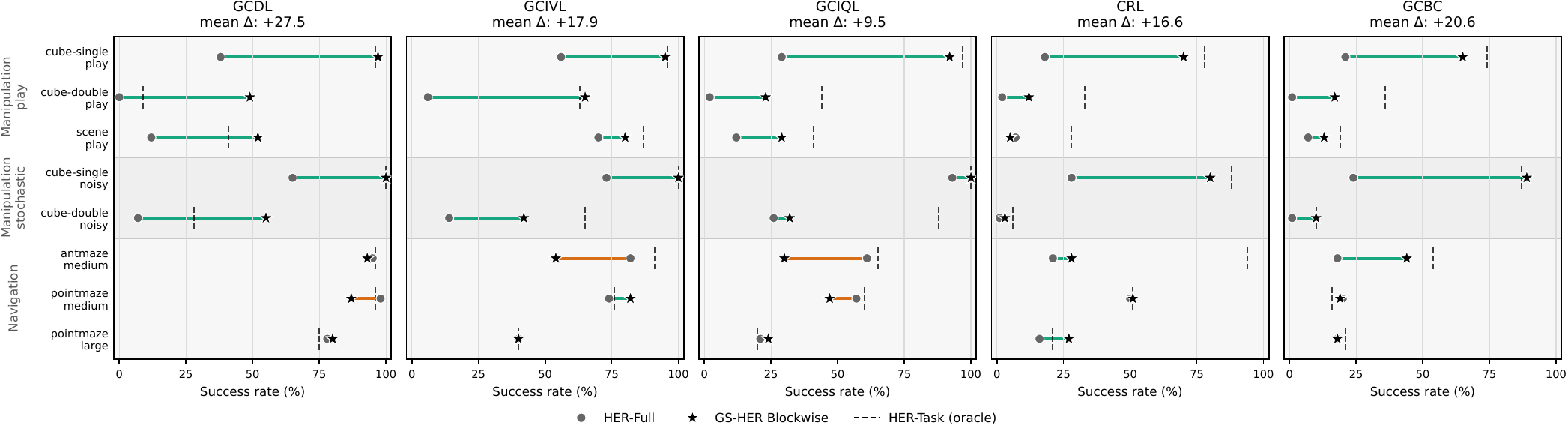}
    \caption{
    \textbf{Full-state hindsight relabeling is bottlenecked by nuisance dimensions.}
    OGBench success rate across base goal-conditioned learners and relabeling schemes. Gray circles denote HER-Full, black stars denote GS-HER Blockwise, and dashed ticks denote the oracle HER-Task projection. Segments connect HER-Full to GS-HER; green indicates improvement and red degradation. Averaged over all manipulation backbone--task pairs, GS-HER improves official success from 24.5 to 55.0, approaching the oracle HER-Task score of 60.6. In contrast, PointMaze acts as a low-nuisance control where full-state and task-state goals are already aligned (47.2/47.5/47.6 for HER-Full/GS-HER/HER-Task).
    }
    \label{fig:official_task_performance}
\end{figure*}

We first evaluate GS-HER on the official OGBench task metric, where success is defined by the benchmark-specific predicate. Fig.~\ref{fig:official_task_performance} shows that GS-HER Blockwise improves over HER-Full on average for every base learner, with the largest gains in manipulation tasks where exact-state matching is misaligned with task success. HER-Task remains a strong oracle reference because it is trained directly on the official task projection, whereas GS-HER is evaluated with the same official query while retaining an inference-time query interface. Thus, query-conditioned relabeling removes much of the full-state bottleneck without changing the underlying offline GCRL objective. Full per-task results are reported in Appendix~\ref{app:main_results_table}.

\noindent
\begin{minipage}[t]{0.48\linewidth}
\vspace{0pt}
\paragraph{Qualitative distance diagnostic.}
To visualize how goal semantics affect the learned value geometry, we inspect the GCDL distance estimator along a successful \texttt{cube-single-noisy-v0} rollout. HER-Full remains saturated near the maximum predicted horizon throughout the episode, even as the cube reaches the goal, because exact full-state matching still requires reproducing nuisance variables such as robot configuration, gripper state, velocities, and object orientation. In contrast, HER-Task decreases with task progress by construction. GS-HER, queried with the official cube-position predicate, closely tracks this task-aligned distance while retaining a single queryable model. Additional plots in Appendix~\ref{app:task_visualizations}.
\end{minipage}
\hfill
\begin{minipage}[t]{0.48\linewidth}
\vspace{0pt}
\centering
\includegraphics[width=0.88\linewidth]{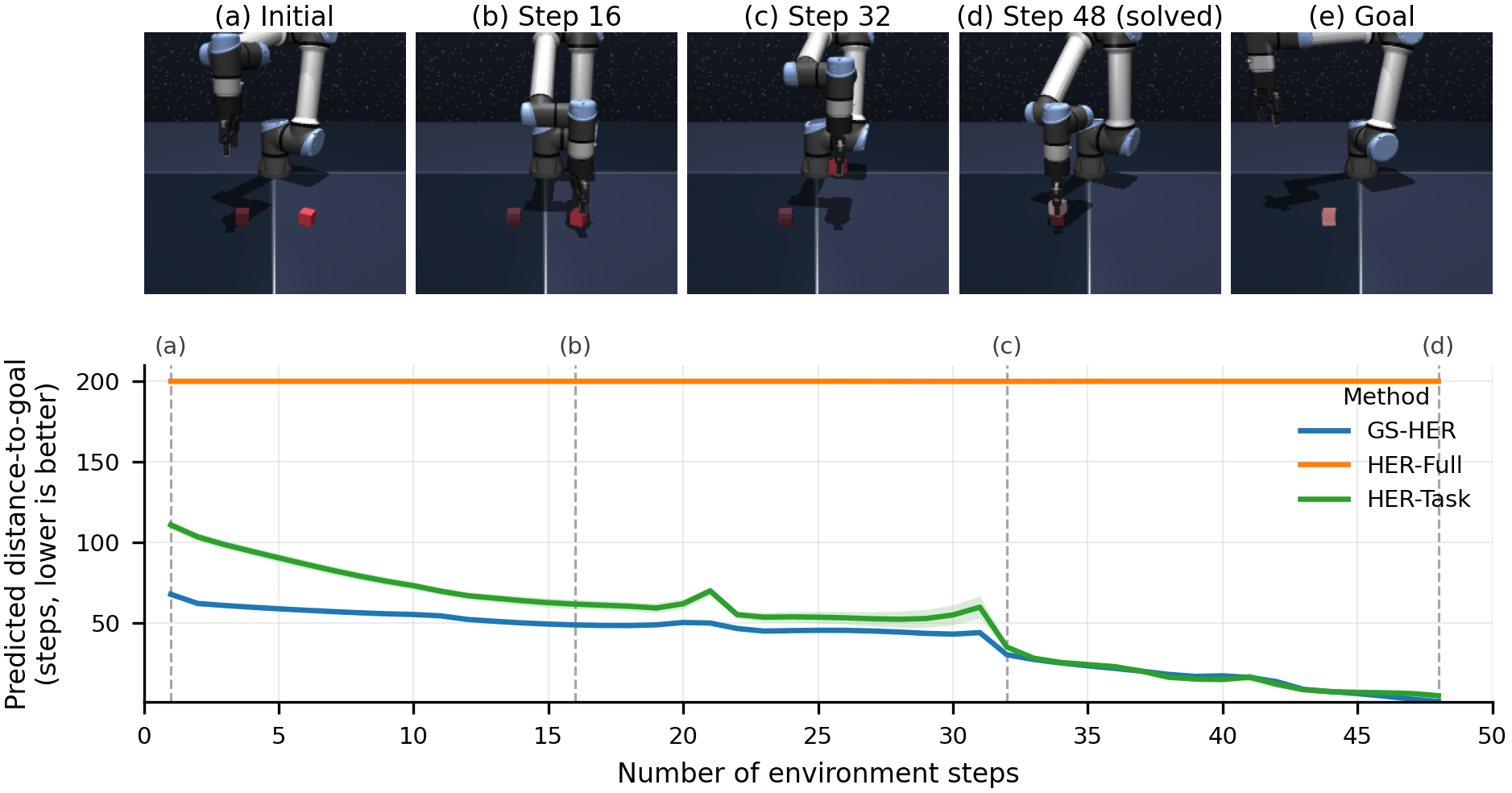}
\captionof{figure}{
\textbf{GS-HER learns task-aligned distance estimates.}
Along a successful \texttt{cube-single-noisy-v0} rollout, GS-HER tracks the oracle task-distance estimate, while HER-Full remains saturated because exact-state goals still depend on nuisance variables beyond the official cube-position predicate.
}
\label{fig:nuisance_analysis}
\vspace{-0.8em}
\end{minipage}

\subsection{One Model, Many Goal Predicates}
\label{sec:exp_one_model_many_goals}
The official benchmark evaluates only one predicate per environment, but this understates the main advantage of query-conditioned relabeling. HER-Task removes nuisance dimensions by fixing an oracle projection before training; changing the predicate therefore requires training a new model. GS-HER instead makes the predicate an input. To isolate this capability from raw official-task performance, we evaluate on \texttt{cube-single-noisy-v0} with GCIQL, where HER-Full, HER-Task, and GS-HER all achieve near-saturated success on the official cube-position predicate. The question is therefore not whether the agent can solve the benchmark task, but whether the same trained checkpoint can be queried with different notions of success.

\begin{figure*}[b]
    \centering
    \includegraphics[width=\textwidth]{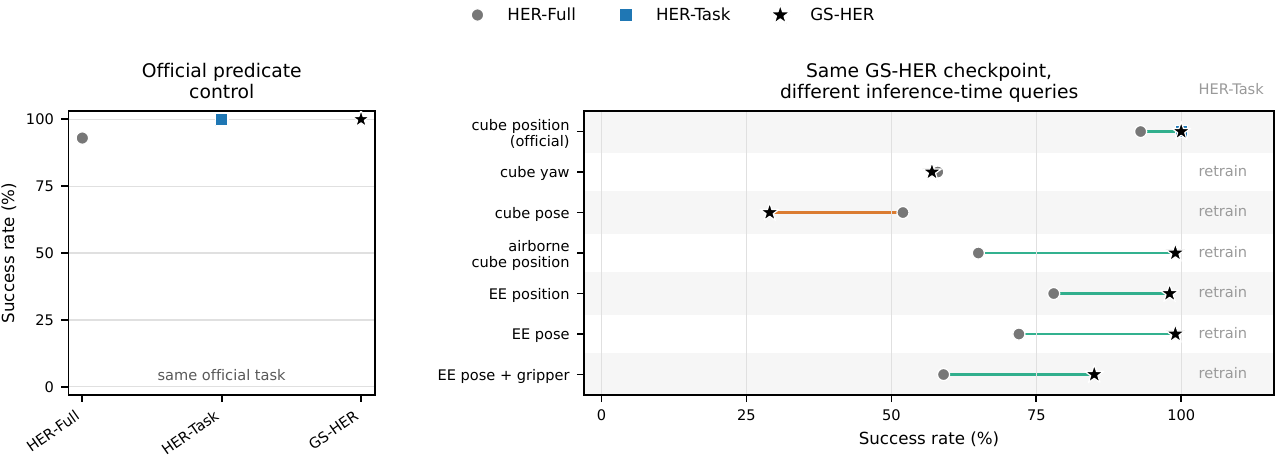}
    \caption{
    \textbf{One model, many goal predicates.}
    A single GS-HER-enabled GCIQL checkpoint is evaluated under multiple inference-time goal queries on \texttt{cube-single-noisy-v0}. Each query activates a different subset of state variables defining success, including object-centric, airborne object, and robot-centric predicates. The checkpoint is fixed across all predicates; only the query changes. The official \texttt{cube\_pos} predicate serves as a control, where HER-Full, HER-Task, and GS-HER all achieve near-saturated success. HER-Task fixes one oracle projection before training and therefore requires retraining to change the predicate, while HER-Full uses a single model but retains exact full-state semantics. Results are computed on a fixed paired suite of held-out validation goals; full predicate definitions and mean $\pm$ standard deviation are reported in Appendix~\ref{app:one_model_many_goals}.
    }
    \label{fig:one_model_many_goals}
    \vspace{-0.8em}
\end{figure*}

Fig.~\ref{fig:one_model_many_goals} shows that a single GS-HER checkpoint can answer a family of goal predicates by changing only the inference-time query. These include the official cube-position predicate, object-centric predicates such as cube yaw and cube pose, an airborne cube-position predicate, and robot-centric end-effector predicates. No model is retrained between predicates. All non-official goals are sampled from a held-out validation split, and the same fixed paired goal suite is reused across methods and seeds. GS-HER preserves perfect official-task performance, remains competitive on cube-yaw control, and strongly improves over HER-Full on airborne and robot-centric predicates, showing that the learned policy can be steered toward substantially different goal semantics from the same checkpoint. At the same time, not every predicate is equally supported: the cube-pose query remains difficult, indicating that queryability does not eliminate differences in predicate complexity or data coverage. This experiment isolates the central capability of GS-HER: HER-Task is an oracle single-predicate baseline, whereas GS-HER exposes a reusable one-model many-goal interface.

\subsection{GS-HER Helps Most When Nuisance Dimensions Dominate}
\label{sec:exp_nuisance_analysis}

\noindent
\begin{minipage}[t]{0.48\linewidth}
\vspace{0pt}
We next ask when query-conditioned relabeling helps most. Counting inactive coordinates alone is insufficient: some coordinates are dynamically coupled to the task, whereas others impose harmful exact-state constraints. We therefore use the gain of HER-Task over HER-Full as an empirical measure of the full-state bottleneck: if an oracle task projection improves over full-state HER, then matching the entire future state is overconstraining the downstream objective.

Fig.~\ref{fig:nuisance_analysis} compares this oracle-projection gain to the gain obtained by GS-HER. Each point is an OGBench task averaged over base learners. GS-HER gains are strongly aligned with the oracle-projection gain, indicating that query-conditioned relabeling helps most in settings where full-state HER is bottlenecked by nuisance dimensions. Points near the diagonal show settings where GS-HER recovers the benefit of the oracle projection without fixing the predicate before training.
\end{minipage}
\hfill
\begin{minipage}[t]{0.48\linewidth}
\vspace{0pt}
\centering
\includegraphics[width=0.88\linewidth]{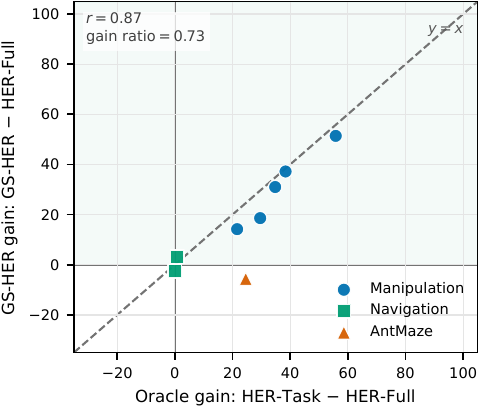}
\captionof{figure}{
\textbf{GS-HER helps when exact-state relabeling is overconstrained.}
GS-HER gains correlate with the oracle-projection gain across OGBench tasks, indicating that query-conditioned relabeling is most useful when full-state HER is bottlenecked by nuisance dimensions.
}
\label{fig:nuisance_analysis}
\vspace{-0.8em}
\end{minipage}

\section{Discussion and Limitations}
GS-HER isolates the relabeling semantics of offline GCRL in state-based settings, where goal predicates can be expressed as coordinate queries. This scope is intentional: it lets us compare HER-Full, HER-Task, and GS-HER under identical learners while separating predicate queryability from perception, language grounding, and real-robot deployment. The results therefore show that making the goal predicate an inference-time input is a useful primitive, not a complete solution to open-ended goal specification.

Extending GS-HER to visual or real-world settings will require replacing coordinate masks with queryable latent factors, object slots, keypoints, or language-grounded state abstractions. Its effectiveness also depends on the query distribution, model capacity, optimization, and offline data coverage; rare or unsupported predicates may not be learned reliably, and mixed-query training may introduce interference.

\section{Conclusion}
\label{sec:conclusion}

We presented GS-HER, a predicate-level generalization of HER enabling queryable goal semantics in offline GCRL. GS-HER allows achieved states to certify entire goal sets, mitigating overconstraint from nuisance variables. Our results on OGBench show that GS-HER enables a single model to answer diverse goal predicates and outperforms standard HER when tasks depend on a subset of state variables. This work highlights the importance of specifying both which state is achieved and which equivalence class it certifies in hindsight relabeling.


\clearpage
\acknowledgments{This work was funded by the Centre for the Development of Industrial Technology (CDTI), Grant No. CER 20251010. ROCCA Project.}


\bibliography{biblio}  

\clearpage
\appendix


\section{Main Results Table}
\label{app:main_results_table}

\begin{table*}[ht]
\centering
\caption{
\textbf{Main OGBench results.}
Average binary success rate (\%) across the five official test-time goals for each task and backbone. We compare full-state HER, oracle task-projected HER, and two GS-HER query distributions: blockwise random queries and semantic-block queries. Results are averaged over 5 training seeds, with standard deviation reported after $\pm$. For each backbone--task row, values within 95\% of the best relabeling strategy are highlighted in bold.}
\label{tab:full_main_results}
\resizebox{\textwidth}{!}{
\begin{tabular}{llcccc}
\toprule
Backbone & Task & \shortstack{HER\\Full} & \shortstack{HER\\Task} & \shortstack{GS-HER\\Blockwise} & \shortstack{GS-HER\\Semantic} \\
\midrule
\multirow{8}{*}{GCDL} & \texttt{cube-single-play-v0} & 38 $\pm$ 3 & \textbf{96 $\pm$ 3} & \textbf{97 $\pm$ 2} & \textbf{96 $\pm$ 1} \\
 & \texttt{cube-single-noisy-v0} & 65 $\pm$ 3 & \textbf{100 $\pm$ 0} & \textbf{100 $\pm$ 0} & \textbf{100 $\pm$ 0} \\
 & \texttt{cube-double-play-v0} & 0 $\pm$ 0 & 9 $\pm$ 1 & \textbf{49 $\pm$ 4} & \textbf{50 $\pm$ 4} \\
 & \texttt{cube-double-noisy-v0} & 7 $\pm$ 1 & 28 $\pm$ 3 & 55 $\pm$ 2 & \textbf{65 $\pm$ 3} \\
 & \texttt{scene-play-v0} & 12 $\pm$ 2 & 41 $\pm$ 3 & 52 $\pm$ 3 & \textbf{58 $\pm$ 2} \\
 & \texttt{pointmaze-medium-navigate-v0} & \textbf{98 $\pm$ 1} & \textbf{96 $\pm$ 2} & 87 $\pm$ 2 & \textbf{98 $\pm$ 1} \\
 & \texttt{pointmaze-large-navigate-v0} & \textbf{78 $\pm$ 2} & 75 $\pm$ 4 & \textbf{80 $\pm$ 3} & 70 $\pm$ 3 \\
 & \texttt{antmaze-medium-navigate-v0} & \textbf{95 $\pm$ 1} & \textbf{96 $\pm$ 2} & \textbf{93 $\pm$ 2} & \textbf{96 $\pm$ 2} \\
\midrule
\multirow{8}{*}{GCIVL} & \texttt{cube-single-play-v0} & 56 $\pm$ 2 & \textbf{96 $\pm$ 3} & \textbf{95 $\pm$ 1} & \textbf{96 $\pm$ 2} \\
 & \texttt{cube-single-noisy-v0} & 73 $\pm$ 2 & \textbf{100 $\pm$ 0} & \textbf{100 $\pm$ 0} & \textbf{100 $\pm$ 0} \\
 & \texttt{cube-double-play-v0} & 6 $\pm$ 1 & \textbf{63 $\pm$ 2} & \textbf{65 $\pm$ 2} & 59 $\pm$ 1 \\
 & \texttt{cube-double-noisy-v0} & 14 $\pm$ 2 & \textbf{65 $\pm$ 3} & 42 $\pm$ 3 & 57 $\pm$ 2 \\
 & \texttt{scene-play-v0} & 70 $\pm$ 3 & \textbf{87 $\pm$ 2} & 80 $\pm$ 1 & 72 $\pm$ 2 \\
 & \texttt{pointmaze-medium-navigate-v0} & 74 $\pm$ 5 & 76 $\pm$ 4 & 82 $\pm$ 2 & \textbf{92 $\pm$ 1} \\
 & \texttt{pointmaze-large-navigate-v0} & \textbf{40 $\pm$ 0} & \textbf{40 $\pm$ 0} & \textbf{40 $\pm$ 0} & \textbf{40 $\pm$ 0} \\
 & \texttt{antmaze-medium-navigate-v0} & 82 $\pm$ 3 & \textbf{91 $\pm$ 2} & 54 $\pm$ 2 & 68 $\pm$ 1 \\
\midrule
\multirow{8}{*}{GCIQL} & \texttt{cube-single-play-v0} & 29 $\pm$ 3 & \textbf{97 $\pm$ 2} & 92 $\pm$ 3 & 92 $\pm$ 1 \\
 & \texttt{cube-single-noisy-v0} & 93 $\pm$ 1 & \textbf{100 $\pm$ 0} & \textbf{100 $\pm$ 0} & \textbf{100 $\pm$ 1} \\
 & \texttt{cube-double-play-v0} & 2 $\pm$ 0 & \textbf{44 $\pm$ 3} & 23 $\pm$ 1 & 16 $\pm$ 2 \\
 & \texttt{cube-double-noisy-v0} & 26 $\pm$ 2 & \textbf{88 $\pm$ 3} & 32 $\pm$ 2 & 27 $\pm$ 3 \\
 & \texttt{scene-play-v0} & 12 $\pm$ 2 & \textbf{41 $\pm$ 2} & 29 $\pm$ 2 & 35 $\pm$ 3 \\
 & \texttt{pointmaze-medium-navigate-v0} & \textbf{57 $\pm$ 3} & \textbf{60 $\pm$ 3} & 47 $\pm$ 2 & \textbf{60 $\pm$ 2} \\
 & \texttt{pointmaze-large-navigate-v0} & 21 $\pm$ 1 & 20 $\pm$ 0 & \textbf{24 $\pm$ 0} & 9 $\pm$ 2 \\
 & \texttt{antmaze-medium-navigate-v0} & 61 $\pm$ 1 & \textbf{65 $\pm$ 2} & 30 $\pm$ 3 & 34 $\pm$ 2 \\
\midrule
\multirow{8}{*}{CRL} & \texttt{cube-single-play-v0} & 18 $\pm$ 2 & \textbf{78 $\pm$ 3} & 70 $\pm$ 2 & 68 $\pm$ 2 \\
 & \texttt{cube-single-noisy-v0} & 28 $\pm$ 3 & \textbf{88 $\pm$ 2} & 80 $\pm$ 2 & 76 $\pm$ 1 \\
 & \texttt{cube-double-play-v0} & 2 $\pm$ 1 & \textbf{33 $\pm$ 2} & 12 $\pm$ 2 & 10 $\pm$ 1 \\
 & \texttt{cube-double-noisy-v0} & 1 $\pm$ 1 & \textbf{6 $\pm$ 1} & 3 $\pm$ 1 & 3 $\pm$ 2 \\
 & \texttt{scene-play-v0} & 7 $\pm$ 2 & \textbf{28 $\pm$ 3} & 5 $\pm$ 2 & 7 $\pm$ 1 \\
 & \texttt{pointmaze-medium-navigate-v0} & \textbf{50 $\pm$ 1} & \textbf{51 $\pm$ 1} & \textbf{51 $\pm$ 2} & 44 $\pm$ 2 \\
 & \texttt{pointmaze-large-navigate-v0} & 16 $\pm$ 5 & 21 $\pm$ 4 & \textbf{27 $\pm$ 4} & \textbf{27 $\pm$ 3} \\
 & \texttt{antmaze-medium-navigate-v0} & 21 $\pm$ 2 & \textbf{94 $\pm$ 1} & 28 $\pm$ 2 & 69 $\pm$ 2 \\
\midrule
\multirow{8}{*}{GCBC} & \texttt{cube-single-play-v0} & 21 $\pm$ 3 & 74 $\pm$ 2 & 65 $\pm$ 2 & \textbf{77 $\pm$ 3} \\
 & \texttt{cube-single-noisy-v0} & 24 $\pm$ 3 & \textbf{87 $\pm$ 2} & \textbf{89 $\pm$ 3} & 78 $\pm$ 3 \\
 & \texttt{cube-double-play-v0} & 1 $\pm$ 0 & \textbf{36 $\pm$ 4} & 17 $\pm$ 3 & 16 $\pm$ 2 \\
 & \texttt{cube-double-noisy-v0} & 1 $\pm$ 1 & \textbf{10 $\pm$ 0} & \textbf{10 $\pm$ 3} & 3 $\pm$ 1 \\
 & \texttt{scene-play-v0} & 7 $\pm$ 1 & \textbf{19 $\pm$ 2} & 13 $\pm$ 2 & 14 $\pm$ 3 \\
 & \texttt{pointmaze-medium-navigate-v0} & \textbf{20 $\pm$ 3} & 16 $\pm$ 3 & \textbf{19 $\pm$ 5} & \textbf{20 $\pm$ 5} \\
 & \texttt{pointmaze-large-navigate-v0} & 18 $\pm$ 4 & 21 $\pm$ 2 & 18 $\pm$ 2 & \textbf{35 $\pm$ 2} \\
 & \texttt{antmaze-medium-navigate-v0} & 18 $\pm$ 2 & 54 $\pm$ 4 & 44 $\pm$ 3 & \textbf{73 $\pm$ 4} \\
\bottomrule
\end{tabular}
}
\end{table*}

\section{Extended Related Works}
\label{app:extended_related_work}

\paragraph{Hindsight relabeling and goal-conditioned value functions.}
Goal-conditioned reinforcement learning learns policies and value functions that generalize across goals. Early work on learning to achieve goals and universal value function approximators formalized this idea by conditioning value functions on both states and goals~\citep{kaelbling1993learning,schaul2015universal}. Hindsight Experience Replay (HER) made sparse-reward goal reaching practical by replaying trajectories with goals that were achieved in hindsight~\citep{andrychowicz2017hindsight}: a trajectory that fails for the commanded goal can still provide successful supervision for another goal that was actually reached.

HER is not inherently restricted to exact full-state goals. Its original formulation allows goals to correspond to predicates or projections of the state, and its manipulation experiments use object-centered goals rather than the complete simulator state~\citep{andrychowicz2017hindsight}. The limitation we study is different: in standard HER, the goal semantics are fixed before training. A policy trained with object-position goals is specialized to object-position reaching; changing the task to object orientation, end-effector pose, gripper state, or a combination of variables requires redefining the goal space and retraining. GS-HER generalizes this setting by making the projection itself query-conditioned. The same achieved state can supervise multiple goal sets, and the same learned policy or value function can be queried at test time with different active variables.

\paragraph{Offline goal-conditioned learning from reward-free data.}
A growing body of work studies goal-conditioned learning from fixed, reward-free datasets. Goal-conditioned behavioral cloning and imitation-style methods imitate actions that lead to future goals~\citep{lynch2020learning,ghosh2020learningreachgoalsiterated,ding2019goal}, while value-based offline GCRL methods use temporal-difference learning, advantage weighting, contrastive objectives, or hierarchical structure to propagate sparse goal-reaching information~\citep{kostrikov2021offline,eysenbach2022contrastive,park2023hiql,park2025ogbench}. Benchmarks such as OGBench formalize this setting by training agents to reach goals sampled from offline trajectories, often using the full state space as the goal space~\citep{park2025ogbench}. This objective is simple and domain-agnostic, but ties learning to exact-state reachability.

Our setting is also related to Actionable Models, which train goal-conditioned Q-functions from fixed robotic datasets using hindsight relabeling, conservative action negatives, and goal chaining~\citep{chebotar2021actionable}. These methods address how to learn stable and useful goal-conditioned value functions from offline data. GS-HER addresses a complementary question: what goal semantics should a hindsight relabeled transition supervise? In standard offline HER pipelines, each achieved state is relabeled under a fixed goal representation, such as a goal image, object-coordinate goal, or full-state goal. In GS-HER, the same achieved state can supervise multiple query-defined goal sets while leaving the underlying offline learner unchanged.

\paragraph{Goal specification and queryable task semantics.}
The choice of goal representation determines what a goal-conditioned agent can be asked to do. Prior work has explored state goals, object-coordinate goals, goal images, automatically generated goals, learned latent goals, language-conditioned policies, and task-conditioned policies~\citep{lynch2020learning,nair2018visual,pong2018temporal,florensa2018automatic,pong2019skew}. These interfaces increase flexibility, but they typically either commit to a fixed success semantics during training or require external task supervision to define the command space. For example, an object-position policy makes success explicit but only for the chosen projection, while a goal-image or latent-goal policy specifies a desired outcome but leaves the relevant invariances implicit in the representation and training objective.

GS-HER exposes goal semantics as an explicit query. A reference state specifies the desired values, while the query specifies which variables define success and which variables are nuisance for the current task. This turns hindsight relabeling from a mechanism for reaching achieved goal states into a mechanism for learning a family of goal-set predicates from the same offline data. Rather than replacing existing goal-conditioned algorithms, GS-HER provides a queryable relabeling interface that can be layered on top of them: the base learner decides how to estimate or optimize reachability, while GS-HER decides what notion of reachability each relabeled transition should supervise.

\paragraph{Hierarchical skills and latent goal spaces.}
GS-HER is related to hierarchical RL and skill-discovery methods that condition behavior on subgoals, latent skills, or temporally extended abstractions~\citep{sutton1999options,bacon2017option,nachum2018data,eysenbach2018diversity,sharma2019dynamics,laskin2022cic}. Some methods discover reusable skills through mutual-information or contrastive objectives~\citep{eysenbach2018diversity,sharma2019dynamics,warde2018unsupervised,laskin2022cic}, while others learn goal-conditioned low-level controllers or hierarchical planners over semantic or learned goal spaces~\citep{levy2017learning,nachum2018data,nachum2018near,co2018self,mendonca2021discovering}. HSD-3 is especially close in spirit because it pretrains low-level skills over variable feature sets and lets a high-level policy select both a goal space and a goal for downstream sparse-reward exploration~\citep{gehring2021hierarchical}. Director extends this line to visual inputs by selecting latent goals inside a learned world model and training a worker to reach them~\citep{hafner2022director}. These works share with GS-HER the view that different decisions may require different active aspects of the state. The role of the goal space, however, is different: in hierarchical methods it defines a temporally extended command or subgoal for online exploration, whereas in GS-HER it defines the success predicate used for offline hindsight relabeling. GS-HER therefore does not introduce a skill hierarchy, world model, or new downstream controller; it is a relabeling operator that can be layered onto existing offline goal-conditioned learners.

In summary, prior work has substantially improved the algorithms, representations, and temporal abstractions used to learn goal-reaching behavior from sparse rewards or offline data. GS-HER is complementary: it changes the semantic object produced by hindsight relabeling, from a single fixed goal representation to a queryable goal set.

\section{Evaluation Tasks Visualization}
\label{app:task_visualizations}

\begin{figure*}[ht]
    \centering
    \includegraphics[width=\textwidth]{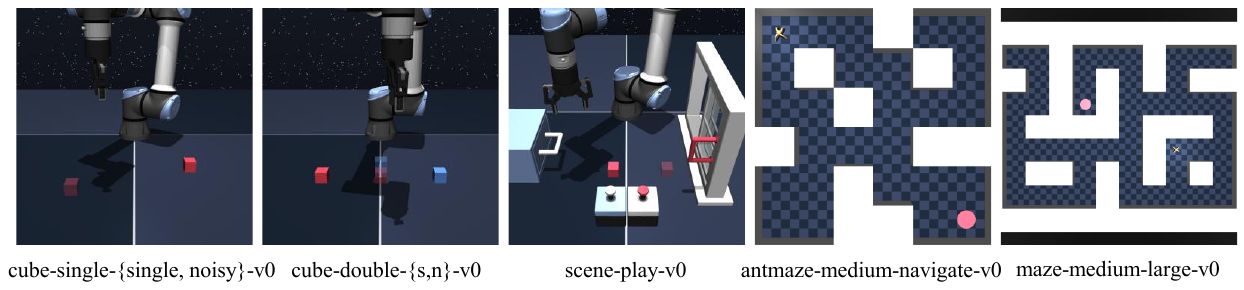}
    \caption{
    \textbf{Evaluation tasks}. State-based OGBench manipulation and navigation domains.}
    \label{fig:tasks}
\end{figure*}

\section{Additional Qualitative Visualizations}
\label{app:qualitative_plots}

\noindent
\begin{minipage}[t]{0.48\linewidth}
\vspace{0pt}
\centering
\includegraphics[width=0.88\linewidth]{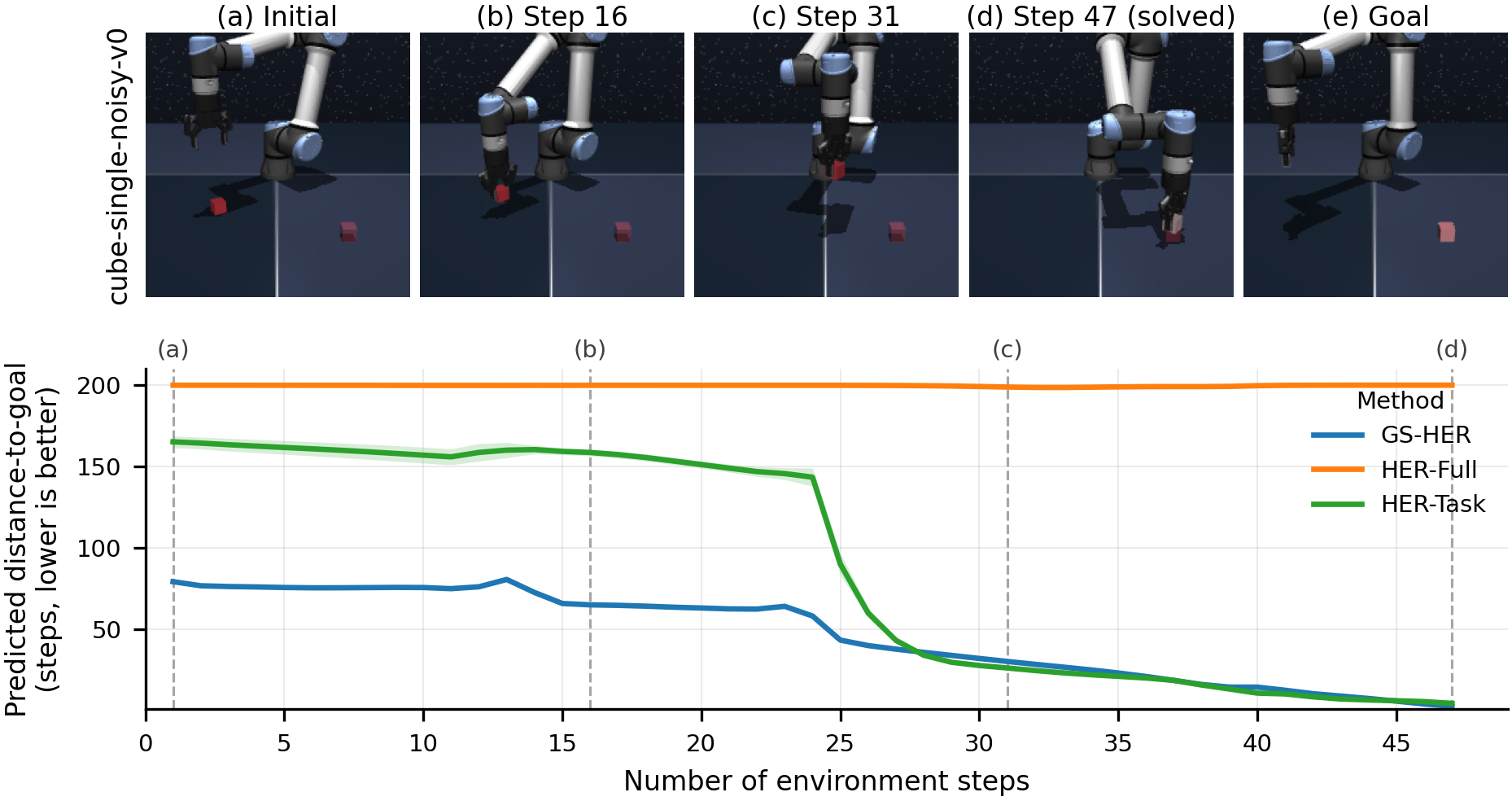}
\end{minipage}
\hfill
\begin{minipage}[t]{0.48\linewidth}
\vspace{0pt}
\centering
\includegraphics[width=0.88\linewidth]{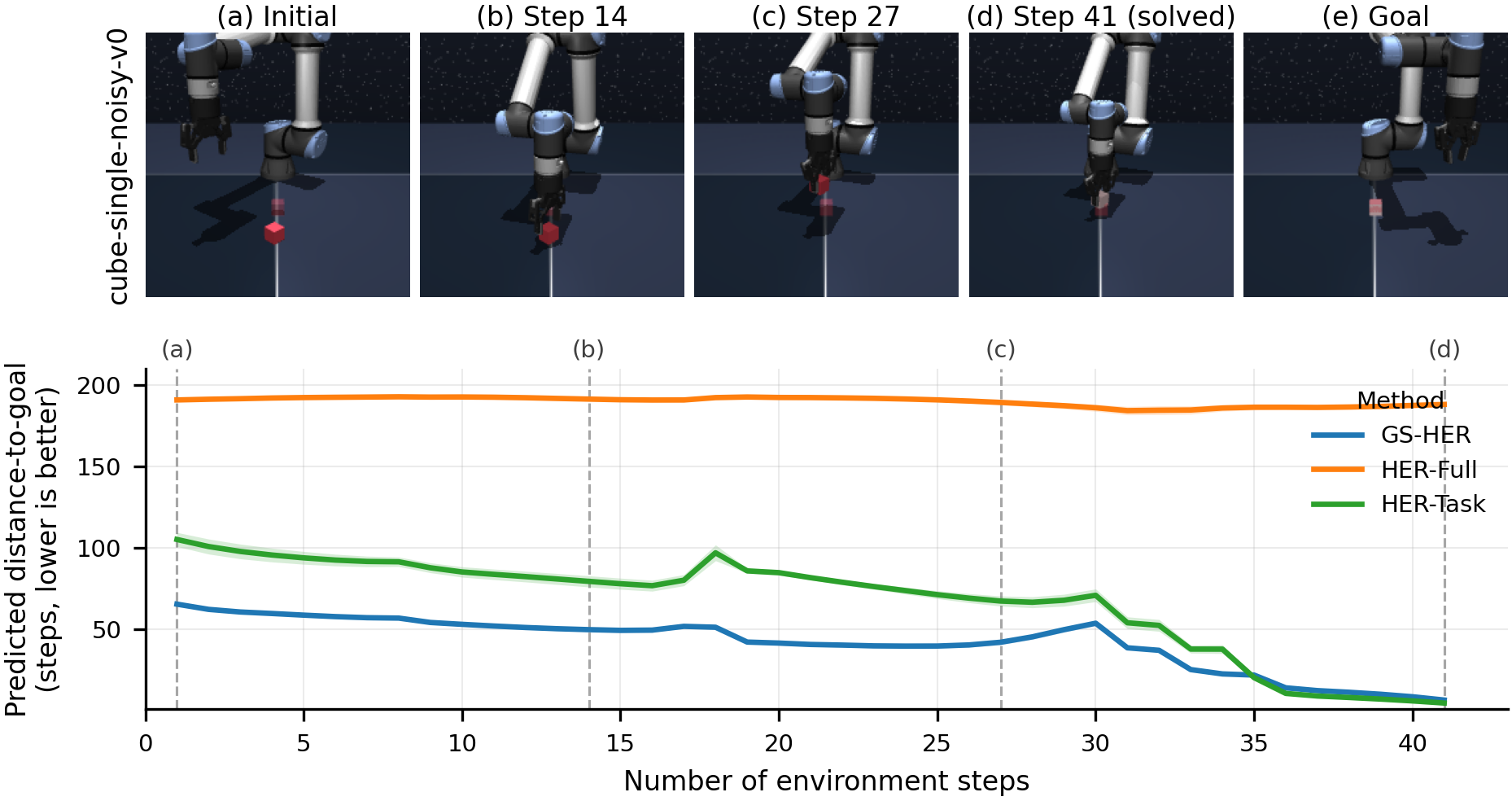}
\end{minipage}

\noindent
\begin{minipage}[t]{0.48\linewidth}
\vspace{0pt}
\centering
\includegraphics[width=0.88\linewidth]{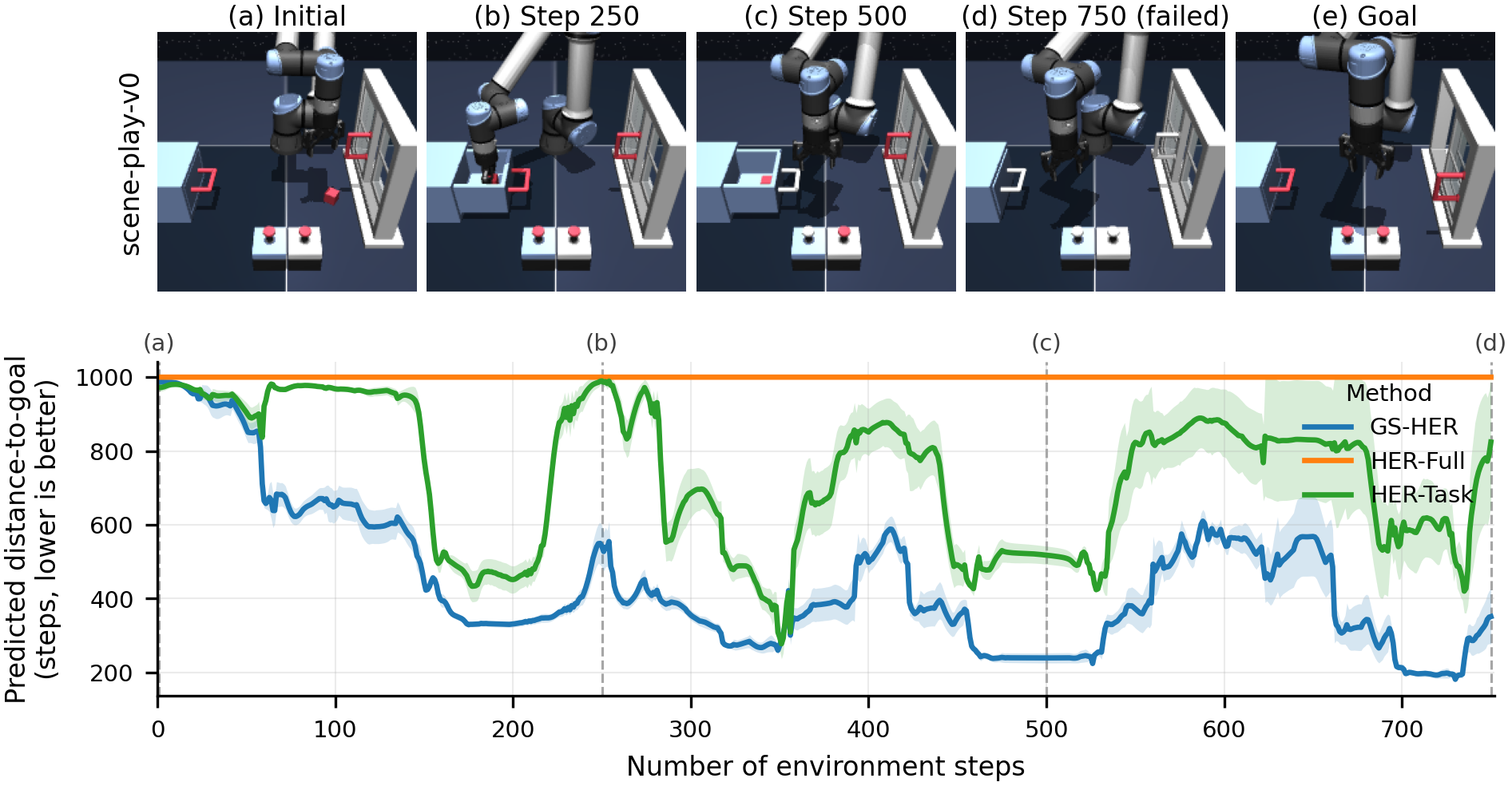}
\end{minipage}
\hfill
\begin{minipage}[t]{0.48\linewidth}
\vspace{0pt}
\centering
\includegraphics[width=0.88\linewidth]{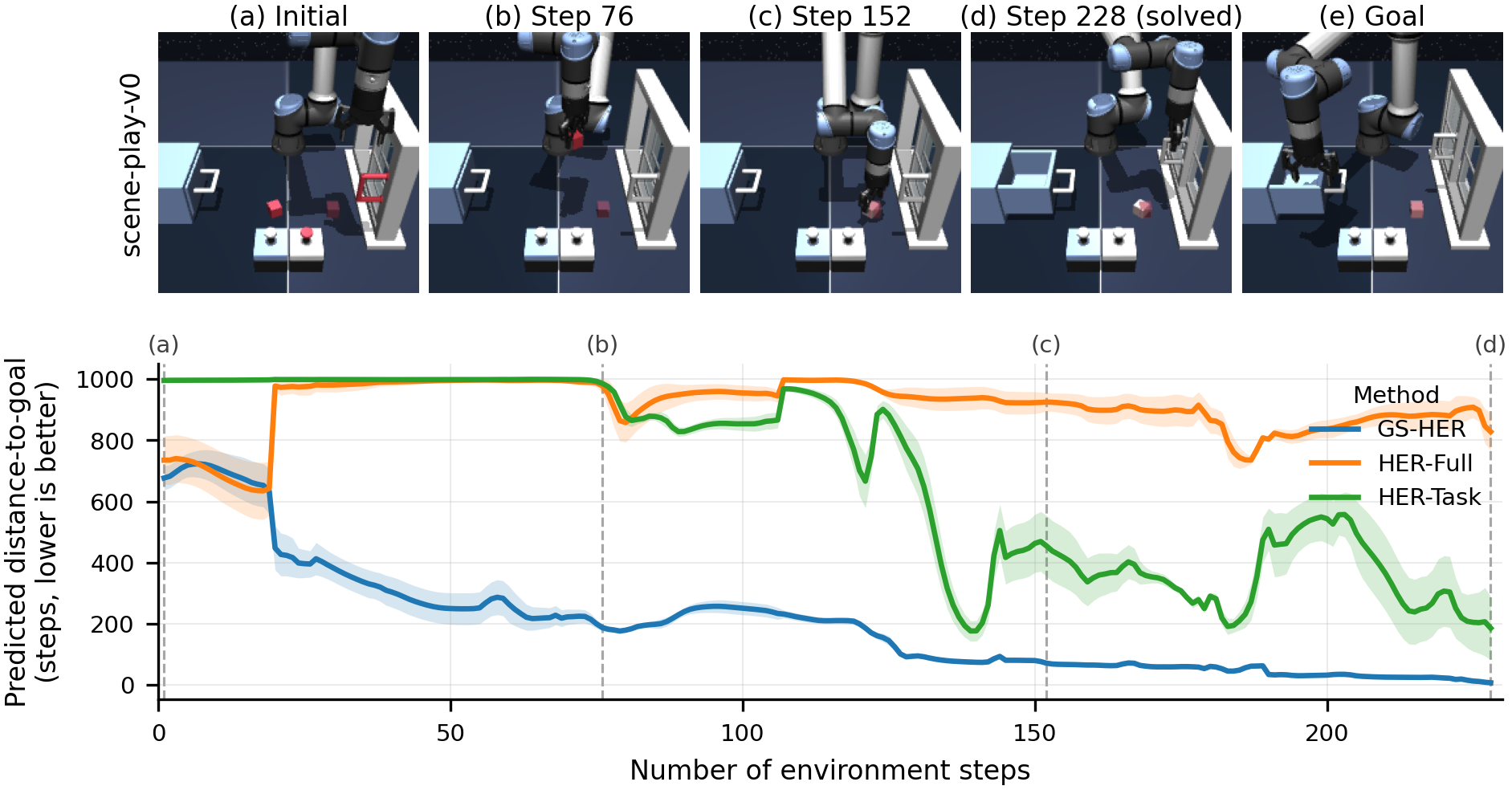}
\end{minipage}

\noindent
\begin{minipage}[t]{0.48\linewidth}
\vspace{0pt}
\centering
\includegraphics[width=0.88\linewidth]{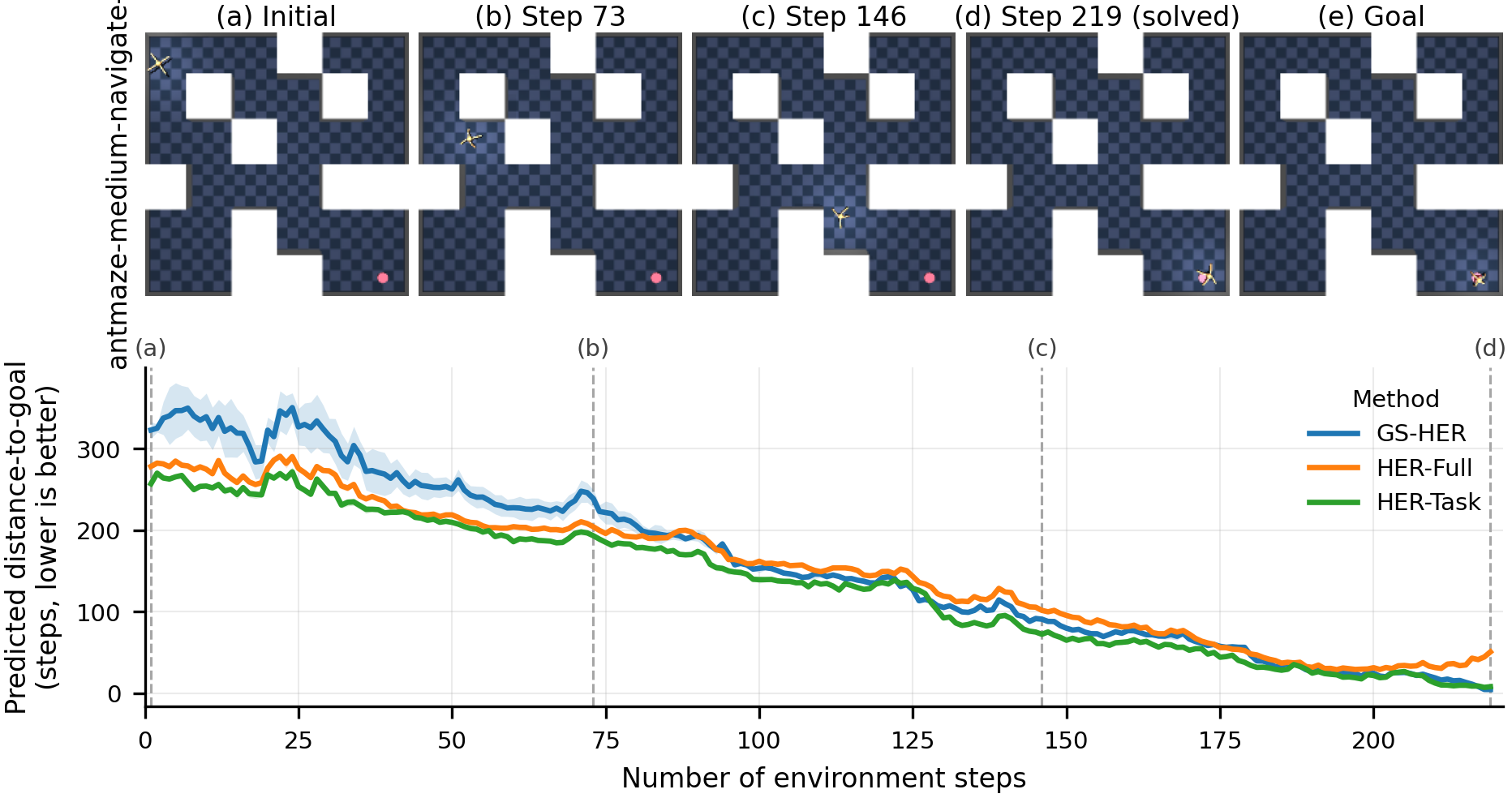}
\end{minipage}
\hfill
\begin{minipage}[t]{0.48\linewidth}
\vspace{0pt}
\centering
\includegraphics[width=0.88\linewidth]{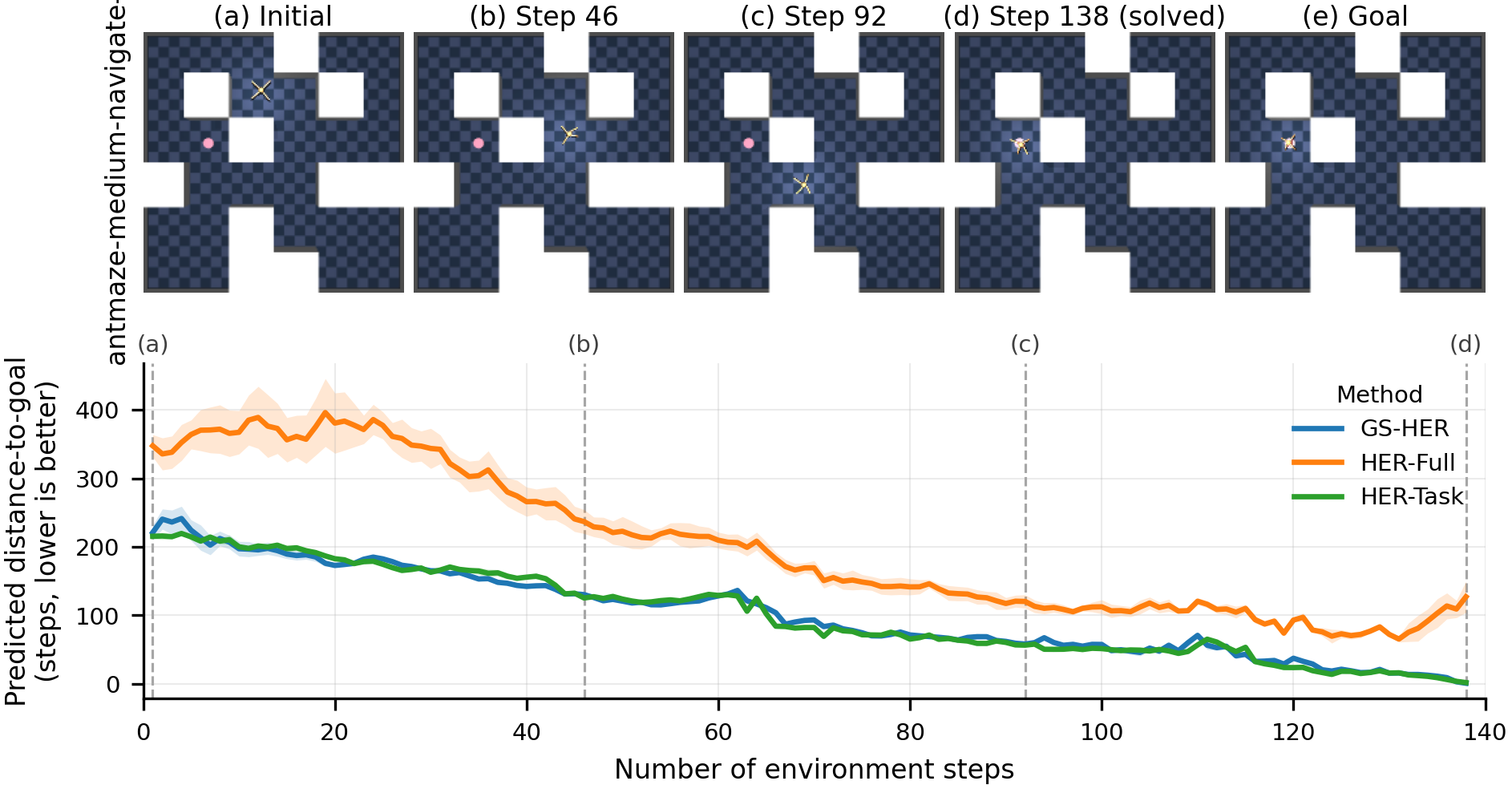}
\end{minipage}

\section{Base Goal-Conditioned Learning Algorithms}
\label{app:base_algorithms}

To isolate the effect of the relabeling strategy, we keep the base goal-conditioned learners as close as possible to the official OGBench implementations~\citep{park2025ogbench}. OGBench provides well-tuned JAX reference implementations and default configurations for representative offline GCRL algorithms, together with detailed algorithmic and implementation descriptions in its appendices. We translate the official JAX implementations to PyTorch. We briefly summarize the base algorithms used in our experiments. We refer readers to OGBench~\citep{park2025ogbench} and the original papers for complete derivations and implementation details.

\paragraph{Goal-conditioned behavioral cloning (GCBC).}
GCBC is the simplest goal-conditioned learner. It trains a policy by supervised imitation, conditioning each state-action pair on a future state from the same trajectory as the goal~\citep{lynch2020learning,ghosh2020learningreachgoalsiterated,park2025ogbench}. Since GCBC does not perform temporal-difference learning, it cannot explicitly propagate reachability beyond demonstrated transitions, but it provides a useful imitation-based baseline for testing whether query-conditioned goal representations can be learned from offline trajectories.

\paragraph{Contrastive reinforcement learning (CRL).}
CRL learns goal-conditioned reachability through a contrastive objective rather than a Bellman backup~\citep{eysenbach2022contrastive,park2025ogbench}. It trains a score function to distinguish future goals from random goals, which can be interpreted as learning a Monte Carlo goal-conditioned value estimate. In OGBench, CRL is a representative contrastive GCRL method and is particularly useful for testing whether GS-HER is compatible with non-Bellman, representation-learning-based objectives.

\paragraph{Goal-conditioned implicit value learning (GCIVL).}
GCIVL is a value-only goal-conditioned variant of implicit Q-learning~\citep{kostrikov2021offline,park2023hiql,park2025ogbench}. It fits a goal-conditioned value function using expectile regression and extracts a policy with advantage-weighted regression. In OGBench, GCIVL is used as a representative value-based offline GCRL method that estimates goal reachability without learning an explicit action-value function.

\paragraph{Goal-conditioned implicit Q-learning (GCIQL).}
GCIQL is the action-value counterpart of GCIVL. It learns both a goal-conditioned value function and a goal-conditioned Q-function using an IQL-style objective~\citep{kostrikov2021offline,park2023hiql,park2025ogbench}. The policy is then extracted using either advantage-weighted regression or a DDPG+BC-style objective, following the OGBench implementation. GCIQL provides a strong value-based baseline, especially in manipulation environments.

\paragraph{Goal-conditioned distance learning (GCDL).}
GCDL is a simple diagnostic distance-learning baseline introduced in this work. It learns a bounded value function whose output can be interpreted as a normalized negative distance-to-goal, and then extracts a policy using advantage-weighted regression. Unlike GCIVL, GCIQL, or CRL, GCDL is not intended as a new competitive offline GCRL algorithm. We include it because its value predictions can be directly rescaled into estimated remaining steps-to-goal, making it useful for visualizing how different relabeling schemes shape the learned value landscape.
\subsection{Goal-Conditioned Distance Learning}
\label{app:gcdl}

We include Goal-Conditioned Distance Learning (GCDL) as a simple diagnostic offline GCRL backbone. GCDL is not a contribution of this work; it is introduced because its value estimate can be interpreted as a predicted remaining distance-to-goal. This makes it useful for visualizing how different relabeling schemes shape the learned value geometry. Importantly, GCDL uses the same unified query-conditioned goal interface as the other backbones in this paper.

\subsubsection{Value Parameterization}
GCDL learns a double goal-conditioned value estimator
\[
    V_{\theta,j}(s,g,q) \in [-1,0],
    \qquad j\in\{1,2\}.
\]
The value is interpreted as a normalized negative distance-to-goal:
\[
    \hat{d}_{\theta,j}(s,g,q)
    =
    -V_{\theta,j}(s,g,q) H_{\max},
\]
where \(H_{\max}\) is the maximum task horizon. Under this convention, \(V_{\theta,j}(s,g,q)=0\) means that the queried goal set has been reached, while \(V_{\theta,j}(s,g,q)=-1\) corresponds to being approximately \(H_{\max}\) steps away. 

\subsubsection{Temporal-Distance Value Learning}

GCDL is trained as an undiscounted shortest-path value estimator with a normalized step cost. Given a relabeled transition \((s_t,a_t,s_{t+1},g,q)\), let
\[
    c_t = c_{g,q}(s_t),
    \qquad
    c_{t+1} = c_{g,q}(s_{t+1}),
\]
and define the normalized one-step cost
\[
    \delta = -\frac{1}{H_{\max}}.
\]
The bootstrapped target is
\[
    y_t =
    \begin{cases}
        0, & \text{if } c_t = 1, \\
        \delta, & \text{if } c_t = 0 \text{ and } c_{t+1}=1, \\
        \delta + \bar{V}(s_{t+1},g,q), & \text{otherwise},
    \end{cases}
\]
where \(\bar{V}\) is computed with an exponential-moving-average target network. With a double value function, we use the pessimistic target estimate
\[
    \bar{V}(s,g,q)
    =
    \min_{j\in\{1,2\}}
    V_{\bar{\theta},j}(s,g,q).
\]
Since values are bounded in \([-1,0]\), the target is clamped:
\[
    y_t \leftarrow \mathrm{clip}(y_t,-1,0).
\]
The value loss is
\[
    \mathcal{L}_V(\theta)
    =
    \mathbb{E}
    \left[
    \sum_{j=1}^{2}
    \left(
    V_{\theta,j}(s_t,g,q) - y_t
    \right)^2
    \right].
\]

\subsubsection{Policy Learning with Advantage-Weighted Regression}

GCDL trains a query-conditioned policy using advantage-weighted regression as GCIVL from~\citep{park2025ogbench}. We compute the advantage in distance space and using the pessimistic value estimate.

\subsubsection{Role in the Paper}

GCDL is intended as an interpretable diagnostic backbone rather than a new algorithmic contribution. Unlike GCIVL, GCIQL, or CRL, whose learned scores are harder to interpret directly, GCDL produces a value estimate that can be rescaled into predicted remaining steps-to-goal. We use this property to visualize how the same \((s,g,q)\) interface induces different value geometries under different query choices. In particular, HER-Full uses \(q=\mathbf{1}\) and therefore estimates distance to an exact full-state goal; HER-Task uses the fixed oracle task query \(q_{\mathrm{task}}\); and GS-HER learns from sampled queries and can be evaluated with the official query at inference time. This allows us to diagnose whether learned distance estimates reflect progress toward the queried goal set or remain sensitive to nuisance variables induced by exact full-state matching.

\section{One-Model Many-Goal Evaluation Details}
\label{app:one_model_many_goals}

This appendix provides the full protocol for the one-model many-goal experiment in
Sec.~\ref{sec:exp_one_model_many_goals}. The purpose of this experiment is to isolate
queryability from raw official-task performance. In the main official benchmark,
\texttt{cube-single-noisy-v0} with GCIQL is already near-saturated for HER-Full,
HER-Task, and GS-HER on the official cube-position task. We therefore use this setting
as a controlled testbed: the question is not whether the agent can solve the official
task, but whether a single trained checkpoint can be queried with different goal
predicates at inference time.

\subsection{Environment and State Coordinates}
\label{app:one_model_env}

We use the state-based \texttt{cube-single-noisy-v0} environment. The observation has
dimension $D=28$ and consists of robot proprioception, end-effector state, gripper
state, and cube state. Table~\ref{tab:one_model_state_layout} gives the state layout
used to construct goal queries.

All predicate rewards are computed in the observation coordinates returned by the
environment, before applying the training normalizer used by the policy. This distinction
matters because the environment internally scales Cartesian positions. Let $x_{\mathrm{raw}}$
denote a raw MuJoCo Cartesian coordinate in meters and $x_{\mathrm{obs}}$ the coordinate
returned in the OGBench state observation. For object and end-effector positions,
the environment uses
\begin{equation}
    x_{\mathrm{obs}} = 10 \, (x_{\mathrm{raw}} - x_{\mathrm{center}}).
\end{equation}
The official cube success criterion uses a raw Euclidean position threshold of
$0.04$m. Thus, the equivalent threshold in observation coordinates is
\begin{equation}
    \epsilon_{\mathrm{pos}} = 10 \times 0.04 = 0.40.
\end{equation}
Yaw variables are represented as $(\cos\theta,\sin\theta)$, so yaw error is measured
by the chord distance between the two unit vectors:
\begin{equation}
    d_{\mathrm{yaw}}(\theta,\theta^\star)
    =
    \left\|
    \begin{bmatrix}
    \cos\theta \\ \sin\theta
    \end{bmatrix}
    -
    \begin{bmatrix}
    \cos\theta^\star \\ \sin\theta^\star
    \end{bmatrix}
    \right\|_2
    =
    2\sin\left(\frac{|\theta-\theta^\star|}{2}\right).
\end{equation}
We use a $20^\circ$ yaw tolerance, corresponding to
\begin{equation}
    \epsilon_{\mathrm{yaw}} = 2\sin(10^\circ) \approx 0.347.
\end{equation}

\begin{table}[h]
\centering
\caption{
State layout for \texttt{cube-single-noisy-v0}.
}
\label{tab:one_model_state_layout}
\resizebox{\linewidth}{!}{
\begin{tabular}{llc}
\toprule
Group & Description & Indices \\
\midrule
\texttt{joint\_pos} & robot joint positions & $0{:}5$ \\
\texttt{joint\_vel} & robot joint velocities & $6{:}11$ \\
\texttt{eff\_pos} & end-effector Cartesian position & $12{:}14$ \\
\texttt{eff\_yaw} & end-effector yaw as $(\cos,\sin)$ & $15{:}16$ \\
\texttt{gripper\_opening} & gripper opening, scaled by the environment & $17$ \\
\texttt{gripper\_contact} & gripper contact signal & $18$ \\
\texttt{block0\_pos} & cube Cartesian position & $19{:}21$ \\
\texttt{block0\_quat} & cube quaternion & $22{:}25$ \\
\texttt{block0\_yaw} & cube yaw as $(\cos,\sin)$ & $26{:}27$ \\
\bottomrule
\end{tabular}
}
\end{table}

\subsection{Goal Predicates}
\label{app:one_model_predicates}

Each evaluation predicate is defined by a reference state $g \in \mathbb{R}^D$ and a
binary query $q \in \{0,1\}^D$. The query specifies which coordinates define success.
Inactive coordinates are nuisance variables for that predicate. A rollout is successful
if the final or any intermediate state satisfies the predicate; success is latched once
achieved, matching the standard sparse-success evaluation convention.

For a predicate with one or more terms, success is defined as the conjunction of all
term-level success tests:
\begin{equation}
    c_{g,q}(s) =
    \prod_{m=1}^{M_q}
    \mathbb{I}\left[d_m(s,g) \leq \epsilon_m\right],
\end{equation}
where $d_m$ is either an $\ell_2$ distance over Cartesian coordinates, an absolute
distance over scalar coordinates, or a chord distance over yaw $(\cos,\sin)$ coordinates.
Table~\ref{tab:one_model_predicates} lists the predicates used in the main experiment.

\begin{table}[h]
\centering
\caption{
Goal predicates used in the one-model many-goal evaluation. All thresholds are in
OGBench observation coordinates before policy normalization. The official cube-position
threshold corresponds to the environment's raw $0.04$m success radius.
}
\label{tab:one_model_predicates}
\resizebox{\linewidth}{!}{
\begin{tabular}{llll}
\toprule
Predicate & Active groups & Metric and threshold & Goal source \\
\midrule
\texttt{cube\_pos}
& \texttt{block0\_pos}
& $\|s_{19:21}-g_{19:21}\|_2 \leq 0.40$
& official env. goal \\

\texttt{cube\_yaw}
& \texttt{block0\_yaw}
& chord distance $\leq 0.347$
& held-out tabletop validation states \\

\texttt{cube\_pose}
& \texttt{block0\_pos}, \texttt{block0\_yaw}
& position $\leq 0.40$, yaw $\leq 0.347$
& held-out tabletop validation states \\

\texttt{airborne\_cube\_pos}
& \texttt{block0\_pos}
& $\|s_{19:21}-g_{19:21}\|_2 \leq 0.30$
& held-out airborne validation states \\

\texttt{ee\_pos}
& \texttt{eff\_pos}
& $\|s_{12:14}-g_{12:14}\|_2 \leq 0.40$
& held-out validation states \\

\texttt{ee\_pose}
& \texttt{eff\_pos}, \texttt{eff\_yaw}
& position $\leq 0.40$, yaw $\leq 0.347$
& held-out validation states \\

\texttt{ee\_pose\_gripper}
& \texttt{eff\_pos}, \texttt{eff\_yaw}, \texttt{gripper\_opening}
& position $\leq 0.40$, yaw $\leq 0.347$, opening $\leq 0.10$
& held-out validation states \\
\bottomrule
\end{tabular}
}
\end{table}

\subsection{Held-Out Goal Suite}
\label{app:one_model_goal_suite}

All non-official reference goals are sampled from a held-out validation split that is
never used to train any evaluated model. The fixed goal suite is generated once and
then reused across all methods, checkpoints, and random seeds. This ensures that
differences in success cannot be attributed to different sampled goal difficulty.

We use a paired evaluation suite consisting of
\[
    7 \text{ predicates} \times 5 \text{ official task ids} \times 20
    \text{ episodes per task} = 700
\]
evaluation episodes. The same list of episode specifications is used for every method.

For the official \texttt{cube\_pos} predicate, we use the environment-provided official
goal. For all other predicates, we sample full reference states from the held-out validation
goal bank. To avoid trivial already-solved goals, sampled goals are rejected if they are
too close to the initial observation under the corresponding predicate distance. We use
a normalized minimum margin of $2.0$ unless otherwise stated.

For object-centric tabletop predicates, we restrict validation goals to states where the
cube is near table height:
\begin{equation}
    0.10 \leq g_{21} \leq 0.35,
\end{equation}
where index $21$ is the cube $z$ coordinate in observation space. This corresponds to
approximately $1$--$3.5$cm in raw height after accounting for the environment's
position scaling. This filter prevents the \texttt{cube\_yaw} and \texttt{cube\_pose}
predicates from being inadvertently dominated by lifted-cube states.

For the airborne predicate, we instead restrict validation goals to lifted states:
\begin{equation}
    g_{21} \geq 0.60,
\end{equation}
corresponding to a raw cube height of at least $0.06$m. In the fixed suite used for our
experiments, the sampled airborne goals satisfy
\[
    \min g_{21}=0.649, \qquad
    \mathrm{mean}\; g_{21}=1.325, \qquad
    \max g_{21}=2.045.
\]
The tabletop predicates satisfy the intended height filter. For example, the generated
suite has
\[
\begin{array}{lccc}
\toprule
\text{Predicate} & \min g_{21} & \mathrm{mean}\;g_{21} & \max g_{21} \\
\midrule
\texttt{cube\_pos} & 0.200 & 0.200 & 0.200 \\
\texttt{cube\_yaw} & 0.190 & 0.212 & 0.322 \\
\texttt{cube\_pose} & 0.180 & 0.210 & 0.316 \\
\texttt{airborne\_cube\_pos} & 0.649 & 1.325 & 2.045 \\
\bottomrule
\end{array}
\]
These statistics are computed before model evaluation and are independent of the evaluated
method.

\subsection{Paired Evaluation Protocol}
\label{app:one_model_paired_protocol}

The evaluation is paired across methods. For each episode specification, every method
is evaluated with the same environment seed, official task id, predicate, and full reference
goal. The environment is reset with the specified \texttt{task\_id} and \texttt{env\_seed}.
The reward returned by the environment is ignored and replaced by the predicate success
defined in Sec.~\ref{app:one_model_predicates}. The original environment dynamics,
observations, action space, and episode horizon are unchanged.

The key comparison in Sec.~\ref{sec:exp_one_model_many_goals} is therefore not that
HER-Task fails on non-official predicates, but that HER-Task does not expose the predicate
as an inference-time input. It is an oracle single-predicate baseline. GS-HER is evaluated
as a single checkpoint whose predicate changes only through the query.

\subsection{Evaluation Metrics}
\label{app:one_model_metrics}

For each predicate and method, we report success rate:
\begin{equation}
    \mathrm{Success} =
    100 \cdot
    \frac{1}{N}
    \sum_{i=1}^{N}
    \mathbb{I}\left[
        \exists t \leq T :
        c_{g_i,q_i}(s_t) = 1
    \right],
\end{equation}
where $N=100$ episodes per predicate in the default suite. Success is latched: once the predicate is satisfied at any timestep, the rollout is counted as successful even if the state later moves away from the goal set.

Across random seeds, we report mean $\pm$ standard deviation over independently trained checkpoints. 

\subsection{Reproducibility Checklist}
\label{app:one_model_reproducibility}

The fixed goal suite is built once before evaluation and saved to disk. During evaluation, every method loads the same fixed suite. The suite is not regenerated
per method or per seed. This is important because some predicates, especially
\texttt{airborne\_cube\_pos} and \texttt{ee\_pose\_gripper}, contain goals that are
substantially harder than others. A paired suite prevents easier or harder sampled goals
from being assigned to different methods.

\subsection{One Model, Many Goal Predicates: Full Results}
\label{app:one_model_many_goals}

Table~\ref{tab:one_model_many_goals_results} reports the final one-model many-goal results on \texttt{cube-single-noisy-v0} with GCIQL. GS-HER is evaluated as a single checkpoint under all predicates, changing only the inference-time query. HER-Full is included as the single-model baseline with exact full-state semantics. HER-Task is reported on the official predicate only, because extending it to the remaining predicates would require retraining a separate oracle-projection model for each predicate. All methods are evaluated on the same fixed paired suite of environment seeds, task ids, predicates, and held-out reference goals.

\begin{table}[h]
\centering
\caption{
One-model many-goal evaluation on \texttt{cube-single-noisy-v0} with GCIQL. All values are success rates, mean $\pm$ standard deviation over training seeds. GS-HER is evaluated as a single checkpoint under all predicates. HER-Full is a single-model exact-state baseline. HER-Task is shown on the official predicate only; evaluating additional predicates would require retraining a separate oracle-projection model for each one.
}
\label{tab:one_model_many_goals_results}
\resizebox{\linewidth}{!}{
\begin{tabular}{lcccl}
\toprule
Predicate & HER-Full & HER-Task & GS-HER Blockwise & Description \\
\midrule
\texttt{cube\_pos}              & 93 $\pm$ 1  & 100 $\pm$ 0 & 100 $\pm$ 0 & official benchmark predicate \\
\texttt{cube\_yaw}              & 58 $\pm$ 3 & retrain & 57 $\pm$ 11 & cube yaw only \\
\texttt{cube\_pose}             & 52 $\pm$ 5 & retrain & 29 $\pm$ 4 & cube position + yaw \\
\texttt{airborne\_cube\_pos}    & 65 $\pm$ 4 & retrain & 99 $\pm$ 1 & lifted cube position \\
\texttt{ee\_pos}                & 78 $\pm$ 4 & retrain & 98 $\pm$ 1 & end-effector position \\
\texttt{ee\_pose}               & 72 $\pm$ 4 & retrain & 99 $\pm$ 1 & end-effector position + yaw \\
\texttt{ee\_pose\_gripper}      & 59 $\pm$ 4 & retrain & 85 $\pm$ 2 & end-effector pose + gripper opening \\
\bottomrule
\end{tabular}
}
\end{table}

\subsection{Interpretation}
\label{app:one_model_interpretation}

The purpose of this experiment is to separate \emph{task performance} from
\emph{queryability}. On the official predicate, all three variants are strong:
HER-Full, HER-Task, and GS-HER all achieve near-saturated cube-position
success. The non-official predicates therefore test a different question:
whether the same trained model exposes a reusable interface for changing the
definition of success at inference time.

HER-Full trains one model, but its implicit predicate is exact full-state
matching. Every coordinate of the reference state is treated as task-relevant,
so HER-Full cannot explicitly express that a cube-yaw task should ignore cube
position, or that an end-effector task should ignore cube state. It can still
succeed on some non-official predicates when satisfying the full reference state
also satisfies the evaluated predicate, but it has no mechanism for declaring
nuisance coordinates irrelevant. HER-Task solves the nuisance problem by fixing
an oracle projection before training, but this produces a model for one
predicate. Changing from cube position to cube yaw, cube pose, end-effector
position, or any other predicate requires redefining the goal space and
retraining a separate oracle-projection model.

GS-HER exposes the predicate as a query. The same trained checkpoint can be
evaluated under object-centric, airborne-object, and robot-centric goal
semantics by changing only \(q\) at inference time. This does not make all
predicates equally easy: compound predicates such as cube pose can remain
limited by data support, exploration coverage, or interference between active
variables. However, the strong performance on airborne cube position and
end-effector predicates shows that the learned policy can be steered toward
qualitatively different notions of success without retraining.

This experiment therefore supports the central claim of GS-HER: it is not
merely a stronger relabeling rule for the official benchmark predicate; it
converts hindsight relabeling into a reusable, queryable goal interface.

\section{Implementation Details}
\label{app:implementation_details}

We implement all offline goal-conditioned learning backbones in PyTorch using a common set of neural modules. This design choice is intended to isolate the effect of the relabeling scheme: HER-Full, HER-Task, and GS-HER differ in the goal representation and success predicate used for relabeling, but not in model capacity or architectural inductive bias. Code implementation available at \url{https://github.com/cvg25/gs_her}.
 
\paragraph{Shared goal-conditioned architecture.}
All policies, value functions, and critics are built from the same goal-conditioned multilayer perceptron, denoted by
\[
    \mathrm{GC\text{-}MLP}_{\theta}(x, g_{\mathrm{in}}),
\]
where \(x\) is the state input, or the state-action input for critics, and \(g_{\mathrm{in}}\) is the goal-conditioning input. The network first embeds \(g_{\mathrm{in}}\) with a shallow MLP and applies layer normalization,
\[
    h_g = \mathrm{LayerNorm}\!\left(f^g_{\theta}(g_{\mathrm{in}})\right).
\]
The goal embedding dimension is set to
\[
    d_{\mathrm{emb}}
    =
    \max\left(8,\left\lfloor d_{g_{\mathrm{in}}}/2 \right\rfloor\right).
\]
The normalized goal embedding is concatenated with the state input and processed by a second MLP:
\[
    y =
    f^{\mathrm{out}}_{\theta}\!\left([x,h_g]\right).
\]
All MLPs use GELU activations. Unless otherwise stated, the hidden dimension is fixed to \(256\) for every backbone and relabeling scheme. We use the same initialization, normalization, and architectural depth across all methods; optimization hyperparameters are reported separately in Appendix~\ref{app:training_details}.

\paragraph{Goal input construction.}
The only architectural input that changes across relabeling schemes is \(g_{\mathrm{in}}\). For full-state HER, the conditioning input is the current state paired with the achieved future state,
\[
    g_{\mathrm{in}}^{\mathrm{HER}}(s,g)
    =
    [s,g].
\]
For HER-Task, we use the fixed task projection \(\phi\) associated with the official benchmark predicate:
\[
    g_{\mathrm{in}}^{\mathrm{Task}}(s,g)
    =
    [\phi(s),\phi(g)].
\]
For GS-HER, the conditioning input is query-conditioned. Given a binary query \(q \in \{0,1\}^D\), we apply it to both the current state and the achieved goal state:
\[
    \bar{s}_q
    =
    q \odot s + (1-q) \odot e_s,
    \qquad
    \bar{g}_q
    =
    q \odot g + (1-q) \odot e_g,
\]
where \(e_s,e_g \in \mathbb{R}^D\) are nuisance-coordinate embeddings for inactive dimensions. The final GS-HER conditioning input is
\[
    g_{\mathrm{in}}^{\mathrm{GS}}(s,g,q)
    =
    [\bar{s}_q,\bar{g}_q,q].
\]
Thus, the query explicitly specifies which coordinates define the current goal predicate, while inactive coordinates are replaced by nuisance embeddings. In our main implementation, \(e_s\) and \(e_g\) are learned parameters. We additionally considered zero and random nuisance embeddings as implementation variants but obtained worse results.

\paragraph{Value functions and critics.}
Scalar value functions are obtained by applying a final scalar head to the shared goal-conditioned MLP:
\[
    V_{\theta}(s,g_{\mathrm{in}})
    =
    \ell\!\left(
    \mathrm{GC\text{-}MLP}_{\theta}(s,g_{\mathrm{in}})
    \right).
\]
When a backbone requires double value estimates, we instantiate two independently parameterized heads with the same architecture. Critics use the same construction, but extend the input with the action:
\[
    Q_{\theta}(s,a,g_{\mathrm{in}})
    =
    V_{\theta}([s,a],g_{\mathrm{in}}).
\]
This ensures that value functions and critics share the same architectural template and differ only in whether the action is included.

\paragraph{Policies.}
Policies are parameterized with the same \(\mathrm{GC\text{-}MLP}\) backbone. The actor predicts the mean of a diagonal Gaussian distribution,
\[
    \mu_{\theta}(s,g_{\mathrm{in}})
    =
    \mathrm{GC\text{-}MLP}_{\theta}(s,g_{\mathrm{in}}),
\]
and defines
\[
    \pi_{\theta}(a \mid s,g_{\mathrm{in}})
    =
    \mathcal{N}\!\left(
        \mu_{\theta}(s,g_{\mathrm{in}}),
        \sigma^2 I
    \right).
\]
The covariance is fixed before temperature scaling. During evaluation, we use a near-deterministic temperature, so action selection corresponds to the predicted mean up to negligible sampling noise.

\paragraph{Backbone reuse.}
The same modules are reused across GCBC, CRL, GCIVL, GCIQL, and GCDL. Imitation-based methods instantiate the actor, value-based methods instantiate the corresponding value and critic modules, and distance- or contrastive-style objectives reuse the same goal-conditioned input interface. Consequently, GS-HER does not introduce a specialized architecture for any backbone. It only changes the relabeled goal semantics and the goal-conditioning input supplied to otherwise identical networks.

\subsection{Training and Benchmarking Protocol}
\label{app:training_details}

Our implementations are written in PyTorch, using the official OGBench JAX implementations as the main algorithmic reference~\citep{park2025ogbench}. Unless otherwise stated, we keep the default OGBench hyperparameters for each backbone and do not perform method-specific hyperparameter sweeps. This choice is important for isolating the effect of the relabeling scheme: HER-Full, HER-Task, and GS-HER use the same optimizer, network capacity, training schedule, and backbone-specific objective hyperparameters.

\paragraph{Single-GPU training budget.}
The original OGBench protocol trains state-based agents for \(10^6\) gradient updates with a minibatch size of \(1024\). To make the full comparison feasible under a single-GPU compute budget, we use a larger minibatch and a shorter optimization schedule. Specifically, we increase the minibatch size from \(1024\) to \(8192\), and scale the learning rate from the OGBench default \(3 \times 10^{-4}\) according to the square-root batch-size scaling rule,
\[
    \eta
    =
    3 \times 10^{-4}
    \sqrt{\frac{8192}{1024}}
    \approx
    8 \times 10^{-4}.
\]
We train each agent for \(2 \times 10^5\) gradient updates. With this schedule, a single run takes approximately \(1.5\) hours on one NVIDIA RTX 5090 GPU, depending on the task and backbone. Thus, our reported results are the OGBench tasks under a reduced-compute Pytorch protocol using the official success metrics.

\paragraph{Evaluation protocol.}
We evaluate each agent at \(160\mathrm{k}\), \(180\mathrm{k}\), and \(200\mathrm{k}\) gradient updates. At each evaluation checkpoint, the agent is evaluated on each of the five official test-time goals with \(20\) rollouts per goal. We report the average success rate across the three evaluation checkpoints. Thus, each reported number averages over
\[
    3 \ \text{checkpoints}
    \times
    5 \ \text{test-time goals}
    \times
    20 \ \text{rollouts}
    =
    300
\]
evaluation episodes per trained agent. This follows the same checkpoint-averaging principle as OGBench, while reducing the number of environment rollouts to fit the available compute budget. 

\paragraph{Reported metrics.}
For every task, backbone, and relabeling scheme, we report the official OGBench goal success rate in percentage. All results are computed over five random seeds, and we report the mean and standard deviation across independently trained agents. For checkpoint-averaged evaluation, the score for each seed is first averaged over the three evaluation checkpoints, and the resulting per-seed scores are then aggregated across the five seeds.

\begin{table}[t]
\centering
\caption{
Training and evaluation protocol used in our experiments.
We keep the default OGBench algorithmic hyperparameters unless explicitly stated.
}
\label{tab:training_protocol}
\begin{tabular}{ll}
\toprule
Item & Value \\
\midrule
Implementation framework & PyTorch \\
Reference implementation & OGBench JAX codebase \\
GPU & NVIDIA RTX 5090 \\
Minibatch size & \(8192\) \\
Learning rate & \(8 \times 10^{-4}\) \\
Gradient updates & \(200\mathrm{k}\) \\
Evaluation checkpoints & \(160\mathrm{k}, 180\mathrm{k}, 200\mathrm{k}\) \\
Rollouts per test-time goal & \(20\) \\
Number of test-time goals & \(5\) \\
Evaluation episodes per checkpoint & \(100\) \\
Evaluation episodes per reported score & \(300\) \\
Hyperparameter sweeps & None \\
\bottomrule
\end{tabular}
\end{table}

\subsection{Data Sampling and Relabeling}
\label{app:data_sampling}

All methods use the same transition sampler. Given a trajectory index \(n\) and timestep \(t\), we sample the transition
\[
    (s_t^{(n)}, a_t^{(n)}, s_{t+1}^{(n)})
\]
and relabel it with a goal state sampled from one of three distributions: the current state, a future state from the same trajectory, or a random state from the full dataset. The probabilities of these three choices are denoted by
\[
    p_{\mathrm{curr}},\qquad
    p_{\mathrm{traj}},\qquad
    p_{\mathrm{rand}},
\]
and are normalized before sampling. The values used in our experiments are reported in Table~\ref{tab:sampling_hyperparameters}.

\begin{table}[t]
\centering
\caption{
Goal-state sampling probabilities used for relabeling.
Value and critic updates use a mixture of current, future-trajectory, and random dataset goals.
Actor updates use only future goals from the same trajectory.
}
\label{tab:sampling_hyperparameters}
\begin{tabular}{lccc}
\toprule
Update type
& \(p_{\mathrm{curr}}\)
& \(p_{\mathrm{traj}}\)
& \(p_{\mathrm{rand}}\) \\
\midrule
Value / critic & \(0.2\) & \(0.5\) & \(0.3\) \\
Actor & \(0.0\) & \(1.0\) & \(0.0\) \\
\bottomrule
\end{tabular}
\end{table}

\paragraph{Current-state goals.}
With probability \(p_{\mathrm{curr}}\), the goal is set to the current state,
\[
    g = s_t.
\]
This produces an immediate positive goal-reaching target and is used by the underlying backbones in the same way as in standard hindsight relabeling pipelines.

\paragraph{Future trajectory goals.}
With probability \(p_{\mathrm{traj}}\), the goal is sampled from a future state in the same trajectory. We sample an offset
\[
    k =
    \left\lfloor
    u^{\alpha} H_{\max}
    \right\rfloor + 1,
    \qquad
    u \sim \mathcal{U}(0,1),
\]
where \(H_{\max}\) is the maximum task horizon used for relabeling. We use \(\alpha=2\), which biases hindsight goals toward nearer future states while still allowing long-horizon relabeling. The selected future index is clipped to the end of the trajectory:
\[
    t_g = \min(t+k, T),
    \qquad
    g = s_{t_g}.
\]
We also store the realized temporal distance \(k_{\mathrm{traj}}=t_g-t\), which is used by backbones that require temporal-distance targets.

\paragraph{Random dataset goals.}
With probability \(p_{\mathrm{rand}}\), the goal is sampled uniformly from the full offline dataset:
\[
    g = s_{\tilde{t}}^{(\tilde{n})},
    \qquad
    \tilde{n} \sim \mathcal{U}\{1,\ldots,N\},
    \qquad
    \tilde{t} \sim \mathcal{U}\{0,\ldots,T\}.
\]
This provides negative or off-trajectory goal supervision for methods that use random goals as part of their objective.

\paragraph{Query sampling for GS-HER.}
For HER-Full and HER-Task, no query is sampled. For GS-HER, each relabeled transition is additionally paired with a binary query
\[
    q \sim p_Q(q),
    \qquad
    q \in \{0,1\}^D,
    \qquad
    \|q\|_0 \geq 1.
\]
We consider different query distributions. The \emph{blockwise} distribution samples contiguous coordinate blocks and requires no semantic state annotations. The \emph{semantic} distribution samples queries over predefined state factors, such as object position, object orientation, robot pose, or gripper state, when such factorization is available. Unless otherwise stated, the main domain-agnostic GS-HER results use the blockwise query distribution, while semantic queries are reported as an oracle-query variant.

The same sampled goal state \(g\) is used for HER-Full, HER-Task, and GS-HER. The only difference is how the goal is represented and how success is evaluated: HER-Full uses the full state, HER-Task uses the fixed task projection, and GS-HER pairs the goal with a sampled query \(q\) defining the active coordinates of the goal set.

\paragraph{Blockwise query sampling.}
The default GS-HER variant uses a blockwise query distribution that exploits only the ordering of the state vector and does not require semantic state annotations. Each query is sampled as one of three types: a full-state query with probability \(0.15\), a random-coordinate query with probability \(0.15\), or a block query otherwise. Full-state queries activate all coordinates and ensure that standard full-state goal reaching remains in the training distribution. Random-coordinate queries activate an unstructured subset of coordinates and improve coverage over arbitrary sparse predicates. Block queries activate the union of one to four independently sampled contiguous coordinate spans. For each span, its length is sampled as a fraction of the state dimension, with scale uniformly drawn from \([0.15,0.40]\), and its start index is sampled uniformly among valid positions. Empty queries are disallowed. This distribution provides a domain-agnostic approximation to structured goal predicates: it encourages the model to answer local coordinate subsets without using privileged semantic labels.





\end{document}